\documentclass[11.5pt]{article}
\usepackage{setspace}
\usepackage{authblk}
\usepackage{booktabs}
\usepackage{rotating,tabularx}
\usepackage[utf8]{inputenc}
\usepackage[margin = 3cm]{geometry}
\usepackage{amsfonts}
\usepackage{amssymb}
\usepackage{amsmath}
\usepackage{amsthm}
\usepackage{xcolor}
\usepackage{hyperref}
\usepackage{algpseudocode}
\usepackage{algorithm}
\usepackage{multicol}

\newcommand{\gran}[1]{G(#1)}

\renewcommand{\emph}{\textbf}

\newcommand{\val}[1]{[\![{#1}]\!]}
\newcommand{\descr}[1]{(\![{#1}]\!)}
\renewcommand{\phi}{\varphi}

\theoremstyle{plain}
\newtheorem{thm}{Theorem}[section]

\theoremstyle{definition}
\newtheorem{definition}[thm]{Definition}

\usepackage{graphicx}

\title{Outlier detection using flexible categorisation and interrogative agendas}
\author[5,6]{Marcel Boersma}
\author[1,6]{Krishna Manoorkar\footnote{Krishna Manoorkar is supported by the NWO grant KIVI.2019.001 awarded to Alessandra Palmigiano.}}
\author[1,2]{Alessandra Palmigiano}
\author[1,6]{Mattia Panettiere}

\author[1]{Apostolos Tzimoulis}
\author[3,4]{Nachoem Wijnberg} 

\affil[1]{School of Business and Economics, Vrije Universiteit, Amsterdam, The Netherlands}
\affil[2]{Department of Mathematics and Applied Mathematics, University of Johannesburg, South Africa}
\affil[3]{College of Business and Economics, University of Johannesburg, South Africa}
\affil[4]{Faculty of Economics and Business, University of Amsterdam, The Netherlands}
\affil[5]{Computational Science Lab, University of Amsterdam, Amsterdam, The Netherlands}
\affil[6]{KPMG, Amstelveen, The Netherlands}

\date{}

\begin{document}

\maketitle

\begin{abstract}
Categorization is one of the basic tasks in machine learning and data analysis. Building on formal concept analysis (FCA), the starting point of the present work is that different ways to categorize a given set of objects exist, which depend on the choice of the sets of features used to classify them, and different such sets of features may yield better or worse categorizations, relative to the task at hand. In their turn, the (a priori) choice of a particular set of features over another might be subjective and express a certain epistemic stance (e.g.~interests, relevance, preferences) of an agent or a group of agents, namely, their {\em interrogative agenda}.  In the present paper, we  represent interrogative agendas as sets of features, and explore and compare different ways  to categorize objects w.r.t.~different sets of features (agendas). 
We first  develop a simple  unsupervised FCA-based algorithm for outlier detection which uses categorizations arising from different agendas. We then  present a supervised meta-learning algorithm to learn suitable (fuzzy) agendas for categorization as sets of features with different weights or masses. We combine this meta-learning  algorithm with the unsupervised outlier detection algorithm  to obtain a supervised outlier detection algorithm. We show that these algorithms perform at par with commonly used algorithms for outlier detection on commonly used  datasets in outlier detection.   These algorithms   provide both local and global explanations of their results. 

\end{abstract}

\section{Introduction} \label{sec:intro}

Outlier detection algorithms serve to identify data points that deviate significantly from the majority of  data in a given context. The deviating data points are referred to as {\em outliers} or {\em anomalies}. Algorithms for outlier detection are  key to a broad range of fields and activities, which include data quality assurance, 
fraud detection, network security,
quality control, and healthcare.

In all these applications, outlier detection helps in improving decision-making, enhancing data reliability, ensuring security, and gaining insights from data that would otherwise be missed. Different algorithms and techniques can be applied depending on the specific domain and the nature of the data.

The present paper focuses on algorithms which identify outliers on the basis of  general {\em categorization principles}.

Categorization or classification of a given data set is one of the central problems in machine learning and data analysis. There is extensive literature on machine learning  techniques and algorithms for clustering, classifying, or categorizing given data for different tasks \cite{jain1988algorithms,jain1999data,murtagh2012algorithms}. However, most of these approaches  lack interpretability or explainability, which makes them not suitable for addressing tasks which require the output to be justified or explained. Application domains  in which the requirement of  explainability  is particularly important include the regulatory and legal domains. 

A natural strategy to make extant classification algorithms  interpretable is to employ  knowledge representation  formalisms, since these are   particularly suitable  for incorporating and representing explanations/justifications given in terms of human knowledge and agency. 
One such important aspects of agency can be formally captured by the notion of \textit{interrogative agenda} (also referred to as {\em research agenda} \cite{enqvist2012modelling}) of an agent (cf.~Section ~\ref{prelim: imterrogative agenda}). Intuitively, given a context, the interrogative agenda of an  agent (or of a group of agents) represents what the agent(s) is (resp.~are collectively) 
 interested in, or what they consider relevant relative to a certain circumstance. 
Within  
the 
formalism of \textit{formal concept analysis} \cite{ganter2012formal}, which is adopted in this paper as a general environment for classification,   the  interrogative agendas of agents are represented as {\em sets of features} of  given {\em formal contexts} (cf.~Section \ref{ssec:FCA}), which are  structures widely used for representing   databases. The effectiveness of a given categorization in a downstream task depends fundamentally on the choice of the particular  agenda used to categorize objects. In \cite{acar2023meta}, a supervised meta-learning algorithm was proposed, which can be used to learn the appropriate (fuzzy) interrogative agendas for classification relative to a given task. This meta-learning algorithm can be used in combination with simple FCA-based algorithms (cf.~\cite{fu2004comparative, prokasheva2013classification,sugiyama2013semi,zhang2014outlier}) for  outlier detection and classification to achieve interpretability for machine learning algorithms  aimed at addressing these tasks.   In this paper, we build on the framework described in \cite{acar2023meta}, by  developing a simple white-box  FCA-based unsupervised algorithm for outlier detection, and combining  this algorithm  with the meta-learning algorithm to obtain an explainable  supervised algorithm for outlier detection. These algorithms can provide both local and global interpretations for their results. We show that these algorithms  can perform at par with the  algorithms commonly used for outlier detection, while also making  their outcomes interpretable.

\paragraph{Background theory.}  The formal framework presented in this paper is set within {\em many-valued Formal Concept Analysis} (FCA) \cite{belohlavek1999fuzzy,wille1996formal}, in which (vague) categories are represented as the formal concepts associated with bipartite weighted graphs seen as {\em fuzzy formal contexts} \cite{belohlavek1999fuzzy,wille1996formal}, each consisting of  domains $A$ (of objects) and $X$ (of features), and a weighted relation $I$ between them. Formal concept analysis  is an influential foundational theory in data analysis 
\cite{ganter2012formal,poelmans2013formal,priss2006formal, valtchev2004formal,   wille1996formal} which provides a framework for categorizing objects w.r.t.~a given set of features.
Mathematically, several important data structures can be described using the framework of FCA. Bipartite graphs can be regarded as instances of formal contexts where two subsets of vertices are taken as  the sets of objects and features respectively, and the set of edges as the incidence relation between them. Binary databases can be regarded as formal contexts with individual entries as objects and the attributes as features. Many-valued databases can be captured  by converting them to discrete databases using conceptual scaling \cite{ganter1989conceptual}. Thus, FCA is a general and uniform method to generate hierarchical and  formally explainable categorizations for objects in several different settings. Furthermore, recent developments in the logical  theory of FCA \cite{conradie2021rough, conradie2017toward, frittella2020toward} lay the groundwork  for addressing and supporting vagueness, epistemic uncertainty, evidential reasoning, and incomplete information. Thus, further developments of the formal framework and algorithms  introduced in this paper will be suitable to address  outlier detection problems for  uncertain, probabilistic or incomplete data.

\paragraph{Formal representation of interrogative agendas.}
The (sub)set of features generating a given  categorization of the objects of  a given formal context  represents the agenda of this categorization, and different agendas will generate different categorizations.  If an agent only considers a certain set of features relevant for a given task, we can model this situation by letting this set of features to be the {\em agenda} of that agent. In many cases, the agenda of an agent  may not be  realistically approximated in such a simple way, but rather, it might be more suitably represented by means  of different relevance or importance scores  assigned to different sets of features. In these cases, agendas will be represented as Dempster-Shafer mass functions, which assign different importance or preference values to different sets of features.  Several methods have been developed in Dempster-Shafer theory  for combining \cite{sentz2002combination} and comparing different mass functions \cite{JOUSSELME2012118}, which allow us to aggregate and compare  the agendas learned by different algorithms and the agendas of human experts. This makes  this framework especially suitable  to formally  study and develop  intelligent systems relying on human-machine interaction.

\paragraph{Outlier detection and explanations.} {In its essence, explainable outlier detection is about combining the task of outlier detection with that of explaining why a certain object has been identified as an outlier. In this context, outliers can be defined in the following, mutually interacting  ways \cite{li2023survey,SEJR2021100172}: (a) by using an {\em oracle}, so that  outliers are identified by an expert (for example, outliers according to labelled test data); (b) by {\em detection}, so that outliers are those elements identified as such by  the detection algorithm;  (c) by {\em explanation}, so that outliers are identified  according to a given methodology, which also allows one to indicate the criteria in terms of which outliers are identified; these criteria can either be indicated as part of the output of the  algorithm itself,  or can be generated by some post-hoc explanation algorithm.}

{Most explanation strategies can be understood in terms of the   alignment of their adopted explanation definition with either an oracle definition, or a detection definition (or both). Choosing one or the other alignment yields fundamentally different types of explainability: indeed, while the first one aims to explain the {\em data}, in the sense that it intends to explain why a certain object  which is declared to be an outlier by the algorithm turns out to be an outlier also  according to some independent definition (e.g.~the one  given by a certain oracle),  the second one aims to explain the {\em procedure} followed by the  algorithm. The post-hoc explanation methods adopted by e.g.~SHAP \cite{Shapley+1953+307+318}, LIME \cite{ribeiro2016should}, COIN \cite{liu2017contextual},  and MINDS \cite{ertoz2004minds}  are  examples of applications of the former strategy, while statistical  models \cite{chandola2009anomaly} such as the KNN-based outlier detection \cite{ramaswamy2000efficient}  or subspace outlier methods \cite{zimek2012survey} are examples of the latter.}

{When  the definition of outlier  given by detection is clear and interpretable, it can be used as an  explanation. The  algorithms  generating this type of outliers are inherently interpretable (e.g.~linear regression, logistic regression, rule-based learners). The unsupervised algorithm introduced in this paper is an example of such algorithm, although the explanations provided by this algorithm may not fully align with a particular oracle definition (e.g.~with the definition according to an expert).  On the other hand, the supervised algorithm introduced in this paper aims at aligning oracle and detection definitions via a learning process which uses the meta-learning algorithm (cf.~Algorithm \ref{alg:meta-learning}).  Specifically, the supervised algorithm learns 
weights to be assigned to different agendas to maximize the alignment between these definitions. 
These weights  represent  the relative importance of the different agendas.  However, as these weights are learned by an iterative learning process, this algorithm may not be regarded as   fully transparent (white-box). 
}

\paragraph{Explainable outlier detection and interrogative agendas.} 
The unsupervised algorithm  introduced in this paper assigns   an FCA-based, {\em outlier degree measure} to any entry w.r.t.~a given agenda. This measure hinges on the simple intuition that the outlier degree of a given object w.r.t.~a given agenda is inversely proportional to the number of other objects which are similar to that object according to that agenda. The final score assigned by the unsupervised algorithm to any object is computed simply  by  taking the mean over the outlier degrees of that object w.r.t.~all agendas.  Thus, this algorithm can be regarded as a white-box  algorithm for outlier detection.  
The final output assigned to any object by the supervised algorithm is computed by taking the {\em weighted} mean of the    outlier degrees of that object w.r.t.~all agendas, where the weights assigned to each agenda are {\em learned} via the meta-learning algorithm (cf.~Algorithm \ref{alg:meta-learning}). These weights  can be used to explain  the  relative contribution of different agendas (sets of features) to the output.  Thus, the approach followed in this paper can be seen as a generalization of the method based on Shapley values \cite{Shapley+1953+307+318} which describe the relevance of individual features in the algorithm results. In this paper, we show that both the unsupervised algorithm and the supervised one perform well on several benchmark datasets for outlier detection, and   that the methodology  adopted in the present paper provides both local and global explanations for their results.

\paragraph{Our contributions.}  
{\begin{itemize}
    \item We propose a novel white-box unsupervised algorithm  based on FCA and interrogative agendas for outlier detection.
    \item We combine this algorithm with a meta-learning algorithm for learning suitable agendas for outlier detection to obtain a supervised algorithm for the same task.
    \item We motivate that these  outlier detection algorithms  are  interpretable/explainable.
    \item We implement these  algorithms on different outlier detection datasets and compare the performance with other commonly used outlier detection algorithms.
    \item We give examples of the explanations provided by the algorithm and suggest how they can be useful in applications. 
\end{itemize}
}
\paragraph{Structure of the paper.} {In Section \ref{sec:preliminaries}, we present the required notation, preliminaries, and  embed the contributions of the present paper in the extant strands of research on FCA-based methods for outlier detection, and related explainability issues and solutions.  In Section \ref{sec:unsupervised}, we introduce a simple FCA-based unsupervised algorithm for outlier detection.  In Section \ref{sec:Learning interrogative agendas}, we describe the framework for learning agendas and provide a generic meta-learning algorithm. We also discuss how this meta-algorithm can be combined with the unsupervised outlier detection algorithm discussed in Section  \ref{sec:unsupervised} to obtain a new supervised algorithm for outlier ddetection. 
In Section \ref{sec:performace}, we describe the performances of our  algorithms on several datasets and compare their performance with the other unsupervised and supervised algorithms. In Section \ref{sec:explnations}, we discuss  the local and global explanability  of our algorithms with examples. In Section \ref{sec:Conclusion}, we conclude and give some directions for future research.}

\paragraph{Implementation.} A fast implementation of the  algorithms discussed above is available at \url{https://osf.io/p4gdy/?view_only=dac7799693d5488ab3061096c04c2f12}\footnote{This is a link to an anonymized version of the repository. The final version of the paper will contain a link to another public git repository.} together with the code used to produce the test results. An online version of the program can be found at \url{https://colab.research.google.com/drive/1kl9l8-2zlkjqMeNpSrMBG9Z3YUe9x4C0?usp=sharing}.

\section{Preliminaries and related work}\label{sec:preliminaries}
Section \ref{ssec:FCA} focuses on basic definitions and facts on Formal Concept Analysis (FCA) and conceptual scaling. Interrogative agendas and their formal representations are discussed in Section \ref{prelim: imterrogative agenda}.  In Section \ref{ssec: related work}, we discuss the most relevant proposals concerning the use of FCA-based techniques for accomplishing outlier detection tasks, and   explainable outlier detection.

\subsection{Formal Concept Analysis}\label{ssec:FCA}
 Developed by Wille \cite{wille1996formal} as a mathematical theory for representing and studying concept hierarchies, Formal Concept Analysis (FCA) has been applied for data-analytic purposes in a broad range of areas which include  linguistics, text retrieval, association rule mining, data analysis \cite{ganter2005formal,ganter1997applied,ganter2012formal,priss2006formal}. 

The basic structures used in FCA as  formal representations of databases are {\em formal contexts} \cite{ganter2012formal}, i.e.~triples $\mathbb{P} = (A, X, I)$ such that $A$ and $X$ are sets of {\em objects} and {\em features}, respectively, and $I\subseteq A\times X$ is an {\em incidence relation} which intuitively reads $a I x$ as `object $a$ has feature $x$' 
for any object $a\in A$ and feature $x\in X$.  
Every formal context as above induces maps $I^{(1)}[-]: \mathcal{P}(A)\to \mathcal{P}(X)$ and $I^{(0)}[-]: \mathcal{P}(X)\to \mathcal{P}(A)$, respectively defined by the assignments 
\begin{equation}
  I^{(1)}[B]: = 
\{x\in X\mid \forall a(a\in B\Rightarrow aIx)\},\quad 
 I^{(0)}[Y] := 
\{a\in A\mid \forall x(x\in Y\Rightarrow aIx)\}.
\end{equation}
A {\em formal concept} of $\mathbb{P}$ is a pair 
$c = (\val{c}, \descr{c})$ such that $\val{c}\subseteq A$, $\descr{c}\subseteq X$, and $I^{(1)}[\val{c}] = \descr{c}$ and $I^{(0)}[\descr{c}] = \val{c}$. 
A subset $B \subseteq A$ (resp.\ $Y\subseteq X$) is  {\em closed} or {\em Galois-stable} if $cl_\mathbb{P}(B) := I^{(0)}[I^{(1)}[B]]=B$ (resp.\ $cl_\mathbb{P}(Y):=I^{(1)}[I^{(0)}[Y]]=Y$).
The set of objects $\val{c}$ of a given concept $c$ is referred to as the {\em extension} (or {\em extent}) of  $c$, while  the set of features $ \descr{c}$ is referred to as its {\em intension} (or {\em intent}). 
The set $L(\mathbb{P})$  of  formal concepts of $\mathbb{P}$  can be partially ordered as follows: for any $c, d\in L(\mathbb{P})$, 
\begin{equation}
c\leq d\quad \mbox{ iff }\quad \val{c}\subseteq \val{d} \quad \mbox{ iff }\quad \descr{d}\subseteq \descr{c}.
\end{equation}
With this order, $L(\mathbb{P})$ is endowed with the structure a complete lattice, the {\em concept lattice} $\mathbb{P}^+$ of $\mathbb{P}$.

{For any $B\subseteq A$, the {\em concept generated by} $B$ is $ (I^{(0)}[I^{(1)}[B]], I^{(1)}[B])$, and likewise, for any $Y\subseteq X$, the {\em concept generated by} $Y$ is $ (I^{(0)}[Y], I^{(1)}[I^{(0)}[Y]])$. A concept is {\em granular} if it is either generated by a singleton subset of $A$ or by a singleton subset of $X$. 
In what follows, $\gran{\mathbb{P}}$ denotes the set of granular concepts of a formal context $\mathbb{P}$. For any formal context $\mathbb{P} =(A,X,I)$ and any $Y \subseteq X$, the {\em relativization} of $\mathbb{P}$ w.r.t.~$Y$ is the formal context $\mathbb{P}_Y:=(A,X,I \cap( X \times Y))$.}

\paragraph{Discretization of continuous attributes and conceptual scaling.} \label{ssec:COnceptual scaling}
{In order to apply FCA on attributes with continuous values, we  need to discretize them.  The process of converting many-valued (possibly continuous-valued) attributes into binary attributes or features for FCA is known as {\em conceptual scaling} \cite{ganter1989conceptual}. Scaling is an important part of most FCA-based techniques and has been studied extensively \cite{ganter1989conceptual,prediger1997logical,prediger1999lattice}. Choosing the correct scaling method depends on the specific task the concept lattice is used for.}

\subsection{Interrogative agendas}
\label{prelim: imterrogative agenda} 

In epistemology and formal philosophy, the {\em interrogative agenda} (or {\em research agenda} \cite{enqvist2012modelling}) of an epistemic agent (or group of agents e.g.,~users) indicates the set of questions they are interested in, or what they want to know relative to a certain circumstance. 
Intuitively, in any given context, interrogative agendas act as cognitive filters that block content which is deemed irrelevant by the corresponding agents. This notion is used to account for the fact that only the information  agents considers relevant is used
e.g.~in the formation of their beliefs, or actions, or decision-making. 
Deliberation and 
negotiation processes can be understood in terms of whether and how decision-makers/negotiators succeed in modifying 
their own interrogative agendas or those of their counterparts, and the outcomes of these processes can be described in terms 
of the  ``common ground'' agenda thus reached.
Also, phenomena such as polarization \cite{myers1976group}, echo chambers \cite{sunstein2001republic} and self-fulfilling prophecies \cite{merton1948self} can be described in terms of the formation and dynamics of interrogative agendas among networks of agents.


Dealing with an outlier detection via classification task, 
different interrogative agendas  lead to the identification of different objects as outliers. For example, if the task consists in identifying outliers in a dataset of sales transactions for a retail business, the outliers identified by an interrogative agenda aimed at  
fraud detection would be very different from those identified by an interrogative agenda aimed at rewarding extraordinary performances of members of an organization. 
Moreover, although a transaction with an exceptionally high value could be an outlier under both the fraud detection and the exceptional performance agendas,  the interpretation and subsequent actions may be very different. 

{For any formal context $\mathbb{P} = (A, X, I)$, we represent the {\em crisp interrogative agenda} of an  agent as a subset $Y$ of $X$. Intuitively, $Y$ represents the set of features the given agent is interested in for  performing a certain outlier detection task. 
Thus, for any formal context $\mathbb{P}=(A,X,I)$,  any crisp agenda $Y$ induces a categorization of the objects in $A$ given by the concept lattice $(\mathbb{P}_Y)^+$ of the relativization of $\mathbb{P}$ w.r.t.~$Y$.  In general, an agent may be interested in different sets of features to a different extent. We account for this situation by means of {\em fuzzy interrogative agendas}. A fuzzy (non-crisp) interrogative agenda is a mass function on $\mathcal{P}(X)$, i.e.~a map $m:\mathcal{P}(X) \to [0,1]$. For  any $Y \subseteq X$, $m(Y)$ represents the importance (or intensity of the preference) of the set of features $Y$ according to the agenda represented by $m$. We assume that the  mass functions are normalized, that is, 
\begin{equation}
\sum_{Y \subseteq X} m(Y)=1.
\end{equation}
 Any such mass function induces a probability or preference function $p_m: \mathcal{R} \to [0,1]$ such that $p_m(\mathbb{P}_Y)= m(Y)$, where  $\mathcal{R} = \{\mathbb{P}_Y \mid Y \subseteq X\}$. 
}
{The agendas of different agents can be aggregated using different Dempster-Shafer rules \cite{denoeux2006cautious, sentz2002combination,shafer1992dempster} to obtain a categorization corresponding to aggregated agendas. A logical framework for deliberation between different agents having different agendas is developed in \cite{FLexiblecat2022}. This framework can be applied to study categorizations when different agents with different interests interact with each other for communication or joint decision-making, as it is the case in several applications like auditing, community analysis, linguistics, etc. }
\label{ssec:interrogativeag}
{\subsection{Related work}}
\label{ssec: related work}
In the present subsection, we discuss some relevant strands of research applying FCA-based methods to outlier detection, and the extant methods for achieving explainable outlier detection.

\paragraph{FCA and outlier detection.}
The extant literature on  outlier detection features several contributions  based on formal concept analysis (FCA) \cite{HU2023110486,10.1007/978-3-319-10061-6_23,sugiyama2013semi,zhang2014outlier}. Sugiyama \cite{10.1007/978-3-319-10061-6_23} suggests  that the outlierness of an object in a concept lattice must  consider the number of concepts it creates. For any formal context $\mathbb{P} = (A, X, I)$,  the outlierness score of a set of objects $B \subseteq A$ proposed by Sugihara is defined as follows:
\begin{equation}
q(B): =1/ |\{ c \in \mathbb{P}^+ \mid B \subseteq \val{c} \, \text{or}\, I^{(1)}[B] \subseteq \descr{c} \}|.
\end{equation}
This definition is geared towards detecting  outliers that belong to a densely agglomerated cluster which locates sparsely w.r.t.~the whole set of  objects. Zhang et al.~\cite{zhang2014outlier} propose an outlier mining algorithm based on  constrained concept lattices to  detect local outliers using a sparsity-based method. 

The FCA-based algorithm for outlier detection developed by  Hu et al.~\cite{HU2023110486} hinges upon the definition of a distance measure between objects  based on granular concepts (cf.~Section \ref{ssec:FCA}) in a concept lattice. This measure is referred to as {\em granular concept-based distinguishing degree} (GCD) and is defined as follows.  For any formal context $\mathbb{P} = (A, X, I)$, and all objects $a, b\in A$,
 \[
 GCD(a,b)= |\{c \in \gran{\mathbb{P}} \mid \text{ either } a \in \val{c}  \text{ and }  b \not\in\val{c}, \text{ or }  a \not\in\val{c} \text{ and }   b \in \val{c}\}|
 \]
 where $ \gran{\mathbb{P}} $ is the set of  granular concepts. Based on this measure,  a {\em distinguishing degree} between  granular concepts   is defined. Namely, For any $c, d\in \gran{\mathbb{P}}$, 
 their {distinguishing degree } is defined as follows: 
 $$M(c,d) : =  \begin{cases}
       \sum_{a \in \val{c},b \in \val{d}} GCD(a,b)/(|\val{c}||\val{d}|) & c\neq d,\\
       0 &  \text{otherwise. }
   \end{cases}   
 $$
 This measure is then used to define the {\em granular outlier degree} (GOD) for  any $c\in \gran{\mathbb{P}}$:
 \[
 GOD(c):= \sum_{d \in \gran{\mathbb{P}}} M(c, d)/|\gran{\mathbb{P}}|.
 \]
 Finally, the {\em granular concept-based outlier factor} (GCOF) of any object $a\in A$ is defined by 
\[
GCOF(a):= \sum_{c\in m_a}GOD(c)W(c)/|m_a|,
\] 

where $m_a=\{c \in \gran{\mathbb{P}} \mid a \in \val{c}\} $, and $W(c)=1-(|\val{c}|/|A|)^{1/3}$. The objects are then ranked according to GCOF scores from largest to smallest, and the higher ranked objects are more likely to be outliers.
 This approach  is similar to the  approach presented in this paper  in that it uses only granular concepts  to improve the scalability of the method. However, there are several key differences between these two approaches.
\begin{itemize}
    \item The first difference concerns the use the present paper makes  of {\em multiple} concept lattices for each formal context, which represent the categorizations induced by different agendas, as   opposed to the use in \cite{HU2023110486} of a {\em single} concept lattice for any formal context, based on a fixed set of features.
    \item The approach presented in this paper only requires us to compute the size of the concepts generated by single data entries, as opposed to computing the concept itself. This makes the algorithm computationally more efficient and scalable.
    \item We assign an outlier degree to an object
    w.r.t.~a concept lattice  directly using a function  the size of the concept  generated by the object in the given  concept lattice.
    On the other hand, the GCOF of an object is calculated through a series of complex computations.  Thus, the outlier degree scores w.r.t.~individual concept lattices according to  our method are more easily understandable to humans. Furthermore, the interrogative agendas framework adopted in the present paper allows us to gauge the importance of different agendas in the output of the algorithm, thus providing a further level of explainability to the algorithm.
\end{itemize}

Finally, it is worth mentioning that the FCA-based approaches using granular concepts for outlier detection may be regarded as a part of the larger class of outlier detection methods based on {\em granular computing} \cite{pedrycz2000granular},  a computing paradigm which processes information in the form of {\em granules}, i.e.~collections of individual information. In FCA, such granulation is provided by the granular concepts (cf.~Section \ref{ssec:FCA}).  Several outlier detection algorithms based on granular computing have been proposed in the literature \cite{chen2008outlier,jiang2015outlier,6845655,YUAN202317}. Most of these approaches \cite{chen2008outlier,jiang2015outlier,6845655} define granules of objects w.r.t.~an attribute  to be the set of objects having the same/similar values for that attribute. An outlier degree measure of an object is then  defined as some function of the size of granules and/or the distance between granules formed from different sets of attributes. The approach which defines outlier degree measures  in terms of   the size of granules  is  similar to the approach, followed in the present paper, which defines the outlier degrees w.r.t.~different agendas (cf.~Section \ref{sec:unsupervised}). However, the  definition proposed here considers granules arising from sets of attributes (agendas) instead of individual attributes. This allows us to detect high level outliers which might not be detected by the analysis w.r.t.~individual attributes (e.g.~the objects with unusual combination of values for certain set of attributes). 

In \cite{YUAN202317}, the significance of an attribute is defined in terms of its {\em fuzzy entropy}, which is a measure of the ability of an attribute to distinguish different objects.  Two sequences of sets of attributes (called {\em forward} and {\em reverse} sequences)
are then constructed by recursively  removing the attributes with the least or the greatest significance from the set of all attributes. Any set of attributes defines a granulation of objects   in which  objects are partitioned into sets having the  same/similar value for all the attributes in that set. Thus, the forward and the backward sequences define a multi-level granulation of the set of objects, which is similar to the granulation obtained by FCA-based methods. The  outlier degree of an object is then defined in terms of  the sizes of the granules generated by an object  and the difference in the sizes of granules at the multiple levels according to forward and reverse sequences. 
This approach is similar to the outlier degree measure assigned  by the unsupervised outlier detection algorithm (cf.~Section \ref{sec:unsupervised}) in this paper, where the space of possible crisp agendas (cf.~Section \ref{ssec:space of possible agendas}) is defined by the sequences of attributes defined  above\footnote{Even though such a choice of space of possible agendas is a viable option, it is different from the choice made in this paper during the evaluation of the  algorithm in Section \ref{sec:performace}.}(even though the exact outlier degrees functions are different).  Moreover, similar to other granular computing-based approaches, the approach in \cite{YUAN202317} also uses a predefined agendas (sequences of sets of attributes) which are given equal importance in the output of the algorithm. In contrast, the meta-learning approach presented in this paper allows us to learn different weights for different agendas.

\paragraph{Explainable outlier detection.}
Research methods in explainable outlier detection  have been 
surveyed in \cite{li2023survey,SEJR2021100172}.  
In particular,  Li et al.~\cite{li2023survey} 
develop  a taxonomy for the various  methods, based on the following six  criteria: (a)  whether the explanation method is applied before ({\em pre-model}) the outlier detection  algorithm, is a {\em part of} the outlier detection algorithm ({\em in-model}), or is applied {\em after}  the outlier detection algorithm ({\em post-model}); (b)  whether the explanations concerns individual outliers ({\em local} explanations), or  the overall model ({\em global} explanations); (c)  whether the explanation method depends on the model (is {\em model-specific}) or is independent from the model  (is {\em model-agnostic}); (d)  whether the explanations are given in terms of features  (are {\em feature-based}), of samples (are {\em sample-based}), or both; (e)  the adoption of some specific explanation technique, such as {\em approximation-based methods} (eg.~LIME \cite{ribeiro2016should}), {\em reconstruction error-based methods} (e.g.~Shapley value-based methods \cite{8973044}), {\em visualization-based methods} \cite{liznerski2021explainable}, {\em intrinsically explainable methods}  (eg.~rule-based methods \cite{he2010co}); (f) the type of  data (eg.~tabular, numerical, categorical). 

As discussed in the introduction, the unsupervised algorithm proposed in this paper  assigns an outlier degree to any object through a simple FCA-based definition  not involving any iterative learning process. Thus, this algorithm satisfies the criteria of  a white-box {\em intrinsically explainable} algorithm according to the  taxonomy discussed above.   On the other hand, the supervised algorithm discussed in Section \ref{ssec:learning algorithm} learns the appropriate weights  of agendas through stochastic descent. Thus, it may not be considered a white-box algorithm. However, it can still explain its results  in terms  of the outlier degree w.r.t.~different agendas and the learned weights (for more  details, see cf.~Section \ref{sec:explnations}). Thus, the supervised  algorithm can still be considered {\em intrinsically explainable} to a high degree. 

As for other criteria in the taxonomy, both the supervised and the unsupervised algorithm provide  in-model, model specific,  local and global, feature-based explanations for tabular data of any  type, e.g.~numerical, categorical data (a  more  detailed discussion about these aspects can be found in Section \ref{sec:explnations}).  The weights of the agendas learned  by the supervised algorithm (cf.~discussion in the introduction)  represent the importance assigned to the different agendas or sets of features by the algorithm. Thus, the role of these weights vis-\`a-vis explainability is similar to that of Shapley-values \cite{Shapley+1953+307+318,8973044}. Shapley values  are commonly used in explainable AI to measure the importance of different features in the output of  a machine learning algorithm. However, the  method presented in this paper differs from the Shapley values in the following ways: (1) The weights of the agendas are learned as a part of the outlier detection algorithm itself. Thus, the  explanation  method adopted here is {\em in-model}, as opposed to Shapley-value based explanations which  are usually {\em post-model}.  
(2) The Shapley value of an attribute in an algorithm is calculated by measuring the effect of removing that attribute on the output of the algorithm.  Hence, the contribution of a given attribute to the output is derived in a counterfactual way, which leaves open and unexplained questions such as how  the interaction between different features influences the outcome (see item (3) below) However, it is unclear how  exactly the removal of some attribute  results in the measured change in the output. In contrast, the output of the supervised algorithm proposed in this paper is the weighted average of the outlier degree w.r.t.~different agendas.  This guarantees a {\em direct}, rather than  counterfactual, dependence of the output on the different agendas. 
(3) Finally, also in connection with the previous point, as the Shapley value  of an attribute is calculated by  measuring the effect of removing  that attribute in the output, it measures the  
significance of this attribute in relation to other attributes in the data. On the other hand,  the output of the supervised algorithm proposed in this paper is the weighted average of the outlier degree w.r.t.~different agendas. Thus, the weight assigned to an agenda represents the   importance of that particular set of features by its own in determining the output.  

 Linear methods like linear regression, logistic regression, and their generalizations like  Generalized Additive Models (GAMs) \cite{chang2023data} have been used in explainable outlier detection. As both the unsupervised and supervised algorithms proposed in this paper  output an overall outlier degree by taking a linear combination of outlier degrees w.r.t.~different agendas, they may be considered as another addition to such generalized linear models for  explainable outlier detection. 

The explainability or the intepretability  of the FCA-based outlier detection methods discussed in the previous paragraph have not been explicitly studied to the best of our knowledge. Sugiyama's method \cite{10.1007/978-3-319-10061-6_23} uses a transparent definition of outlier, and thus can be considered a white-box algorithm. However, this method lacks scalability due to its heavy computational requirements.  The method proposed by  Zhang et al.~\cite{zhang2014outlier} can possibly  provide the explanations in terms of sparsity of certain subspaces based on concept lattices. However, the explainability  aspect of this method has not been studied yet, and needs further exploration. In the previous paragraph, we have already remarked that the outlier degree measure (GCOF) used in  \cite{HU2023110486} is a complex definition which is the result of a series of several other computations, and as such is not easily parsed and interpreted by   a human.  Note that all  these approaches use a single concept lattice for outlier detection as opposed to the interrogative agenda based framework used in this paper, which uses a  combination of several different concept lattices based on different agendas. This  allows us to  interpretat the results in terms of the conceptual epistemic framework of interrogative agendas.

\section{Unsupervised outlier detection algorithm using FCA and interrogative agendas}\label{sec:unsupervised}
In this section, we describe a simple unsupervised algorithm for outlier detection using FCA. This algorithm is based on the idea that, for any given formal context $\mathbb{P} = (A, X, I)$ and any object $a\in A$, the closure of $a$ in $\mathbb{P}$ (cf.~Section \ref{ssec:FCA}), which we denote $cl_\mathbb{P}(a)$ or $cl(a)$ if $\mathbb{P}$ is clear from the context, represents the set of objects which are ``similar'' or ``almost indistinguishable from $a$''. An object is more likely to be an outlier if it has fewer objects similar to it; thus, the outlier degree of $a$ w.r.t.\ some formal context is defined in terms of the size of its closure as follows: 

\begin{definition}\label{def:outdeg(a,Y)}
For any formal context $\mathbb{P}=(A,X,I)$, and any $a \in A$, the {\em outlier degree} of $a$ with respect to $\mathbb{P}$ (or according to $\mathbb{P}$) is $outdeg(a,\mathbb{P}^+) := \exp{(-\gamma*|cl_\mathbb{P}(a)|)^2}$, for some parameter $\gamma \in \mathbb{R}$.
\end{definition}
The choice of the parameter $\gamma$ depends on the given dataset and number of bins\footnote{\label{footn:gamma}In the evaluation of the algorithms performance in this paper (cf.~Section \ref{sec:performace}), we have chosen  parameter $\gamma$ randomly in the interval $[0,1]$. In the future, we intend to further study the optimization of $\gamma$ for the best performance.}. Based on this measure, we can obtain a simple unsupervised algorithm for outlier detection which,  for any formal context $\mathbb{P} = (A, X, I)$, assigns  the following outlier degree to any object $a\in A$ and any collection $\mathcal{T}\subseteq \mathcal{P}(X)$ of chosen crisp agendas:   %

\[
outdeg(a, \mathcal{T}) = \sum_{Y\in \mathcal{T}} outdeg(a, \mathbb{P}_{Y}^+)/|\mathcal{T}|. 
\]
 That is, for a given collection $\mathcal{T}$ of agendas, the outlier degree assigned to a given object $a$ by this algorithm 
 is the {\em average} of the outlier degrees of $a$ relative to each agenda in $\mathcal{T}$. 
The performance of this simple algorithm is detailed in Table \ref{tab:unsupervised_results}, Section \ref{sec:performace}, and is compared with other extant outlier detection algorithms. In Section \ref{sec:Learning interrogative agendas}, we define a supervised variant of this algorithm, in which different weights for the outlier degrees according to different contexts are learnt through an iterative process.

\paragraph{Computing closures.}
\label{ssec:computing_closures}
 Efficiently implementing  the algorithm  discussed above requires the efficient computation of the outlier degree of a given object w.r.t.~a given lattice. Since the outlier degree of an object $a$ in a formal context $\mathbb{P} = (A, X, I)$ w.r.t.~an agenda $Y$ is determined by the size of the closure of $a$ in the concept lattice $\mathbb{P}_Y^+$,  {the computation of the outlier degree of $a$} boils down to the efficient computation of the sizes of the closures of  objects in any formal context. In general, computing the concept lattice of a given formal context $\mathbb{P} = (A, X, I)$ is  done in exponential time, with polynomial delay. That is, in the state of the art algorithms \cite{fcalineardelay}, it is possible to enumerate the Galois-stable sets with polynomial delay time, i.e., the time to get the next element in the enumeration is $\mathcal{O}(|A||X|)$. Computing the closure of a set of objects or features requires to compute two intersections (the intersection of the extents of the attributes which are shared by all the objects), but computing the size of the  closure of  an object (a singleton set) only requires to compute  one intersection, i.e., the intersection of the extents of the features possessed by the object.  Practically, the computation of this intersection can   be executed  efficiently if the formal context is stored twice, i.e., by storing for each object (resp.\ feature) encode its intent (resp.\ extent) in a bit vector indicating for each feature (resp.\ object) whether the object (resp.\ feature) has it. This encoding allows us to compute the outlier degrees of all objects in $\mathbb{P}$   in  $\mathcal{O}(|A||I|)$ in a straightforward manner. Unlike most FCA-based algorithms for outlier detection (cf.~Section \ref{ssec: related work}), the algorithm presented above does not need computing the whole concept lattice (or a part thereof); it only needs computing {\em closures of singletons}. This design feature  makes the algorithm above more scalable than those  discussed in Section \ref{ssec: related work}.

\section{Learning interrogative agendas} \label{sec:Learning interrogative agendas}
{
In this section, we recall and further develop the FCA-based framework for learning agendas introduced in \cite{acar2023meta}.
As discussed in Section \ref{sec:intro},  the degree of effectiveness in accomplishing a certain outlier detection task  via FCA-based categorization depends on the choice of the set of features used to categorize the given set of objects. 
However, for many applications, it is hard to estimate which features are of importance, and how important they are relative to the specific task at hand; that is, it is difficult to identify the optimal agenda for a given task. In the present section, we describe a  framework that addresses this problem by using a {\em meta-learning}   algorithm (i.e.~Algorithm \ref{alg:meta-learning}),   which aims at identifying the appropriate agenda for a given task. { As discussed in Section \ref{sec:performace}, this meta-learning algorithm can be used in combination with FCA-based    outlier detection or classification algorithms, and will improve their performances   by choosing the appropriate (fuzzy) agenda based on which each given algorithms can perform its task.}}
\subsection{The space of possible agendas}\label{ssec:space of possible agendas}
Let $\mathbb{P}=(A,X,I)$ be a formal context and let 
$\mathcal{R} = \{\mathbb{P}_Y \mid Y \subseteq X\}$ be the set of all its relativized contexts. 
As discussed in Section \ref{prelim: imterrogative agenda}, a (non-crisp) interrogative agenda on a given set of features $X$ is given by a mass function $m:\mathcal{P}(X) \to [0,1]$, where the value $m(Y)$ denotes  the relevance assigned to the agenda $Y\subseteq X$  relative to the given  task. 
Thus, the set of all (non-crisp) agendas associated with $\mathbb{P}$ can be identified with the set of probability functions on $\mathcal{R}$, or equivalently, with the set of mass functions  on $\mathcal{P}(X)$. As
discussed in the introduction, we want to learn which fuzzy  agenda  leads to an optimal categorization  relative to the effective accomplishment of a given task. This task corresponds to learning a mass function $f$ on $\mathcal{R}$ which represents a suitable categorization for the given task. That is, we use machine learning to
search for a “good” function in the space of mass functions on $\mathcal{R}$.  
For the sake of computational and notational convenience,  we propose the following simplifications.

Let  $\mathbb{R}$ be the set of real numbers.  We represent {\em weight functions} either  as maps $f:\mathcal{R} \to \mathbb{R}$ or equivalently, as maps $w: \mathcal{P}(X) \to \mathbb{R}$. 
Intuitively, for any agenda $Y$, the value  $f(\mathbb{P}_Y)$ (resp.~$ w(Y)$) denotes the importance (or preference) assigned to $Y$.   Any such function $f$ (resp.~$w$) can be seen as a real-valued vector of dimension $|\mathcal{R}|$ (i.e.~$|\mathcal{P}(X)|$). Thus, the set of all such functions is isomorphic to the space $\mathbb{R}^{|\mathcal{R}|}$.  As this space is linear, the shift from  mass functions on $\mathcal{R}$ to  the real-valued functions simplifies the task of learning an (fuzzy) agenda  (via the identification of a suitable weight function $f$) that minimizes a given loss function using a simple gradient descent method. 

The weights  assigned to the formal contexts  can be interpreted as probabilities on $\mathcal{R}$, (and hence  mass functions on $\mathcal{P}(X)$) via normalization when all the weights are non-negative. The negative weight assigned to a (crisp) agenda suggests that the categorization  induced by that particular agenda is opposite to the preferred categorization for the task at hand. For example, suppose we are interested in detecting elements  with a value  of feature $x$ being abnormally  high, while the outlier detection  algorithm to which the meta-learning algorithm is applied finds 
outliers with value  of $x$ abnormally  low. Then the meta-learning algorithm is likely to assign a negative weight to the  agenda $\{x\}$. 

As discussed earlier, one major problem in applying FCA-based methods for outlier detection or classification  is the  computational complexity of the resulting algorithms.  The proposal of the present paper is to consider priority (or weight) functions on a set of  concept lattices corresponding to different agendas. The fact that the number of different (crisp) agendas arising from a set $X$ of features is exponential in $|X|$ may add another exponential factor to the complexity of the algorithm, which can make the problem computationally unfeasible in many cases in which the number of features is large. Thus, in most applications, we  need to restrict the domain of definition of the weight functions to a smaller 
subset  $\mathcal{T}\subseteq \mathcal{P}(X)$ of {\em designated agendas}. 
We propose the following strategies for choosing this restriction. 

\begin{enumerate}
    \item \textbf{Choosing small agendas.} This strategy consists in fixing some   $\alpha\ll |X|$, and letting the collection of designated  agendas be $\mathcal{T}: = \{Y\subseteq X\mid |Y|\leq \alpha\}$. 
    This strategy is based on the idea that   classification or outlier detection tasks can be performed with good accuracy by considering only a small number of features together. This is especially the case of tasks involving epistemic dimensions, given that  humans use  only a limited number of features in combination for  comparison and classification tasks.
     Agendas with a small number of features are also easier to interpret for humans.    In some situations, it might also be useful to add the {\em full agenda}, i.e.~the agenda given by set $X$ of all features,   to the collection $\mathcal{T}$. This allows the algorithm  to weigh in  the full concept lattice with all available information for the task at hand, while allowing  the possibility of giving a higher or lower (compared to other features) importance to some small subsets of features. Consider a profiling scenario where a small number of  suspects are identified as outliers, within a large population, based on a  set of features  $X$. {Let $x \in X$ be the 
     feature `previous conviction for violent crime'. Suppose $\alpha$ is set to $1$, and  it is known that  feature $x$ has much larger importance in identifying potential suspects compared to all other features in $X$.  Then the appropriate weight function learned in this scenario will assign much higher weight  to agenda $\{x\}$ and possibly agenda $X$ (in case the full agenda is also significantly important for outlier detection)   compared to
     all other agendas. 
     }
    
    
    \item  \textbf{Choosing designated  agendas based on prior or expert knowledge.} This strategy consists in using  prior or expert knowledge, when available, for deciding which agendas  are particularly  important or relevant for certain tasks. In such cases, these agendas can be  taken as the set of basis set of agendas for meta-learning.  In addition, this strategy makes it possible  to incorporate prior or expert knowledge into  FCA-based algorithms. 
    \item  \textbf{Choosing designated agendas adaptively.} In applying this strategy, we start with a collection of agendas consisting of all  subsets $Y\subseteq X$  s.t.~$|Y|<\alpha$  for some small $\alpha$ (usually taken as 1). We use   Algorithm  \ref{alg:meta-learning}
    to assign weights  to them, and then drop all the agendas which get assigned a very low weight (after normalization).  We then consider agendas $Y$ which are  subsets of the union of the agendas that are not removed in the first step and such that $|Y|<\alpha_1$ for some small $\alpha_1 > \alpha$. Choosing  these agendas can be interpreted as considering combinations of features that are deemed important in the first learning step.  We then repeat the learning process with this new set of agendas. We  iterate this process until all the agendas added in the last step get assigned low weights or we reach  the full agenda $X$. In this way, at each step, we recursively check the possible combinations of  agendas  deemed to be important so far in the next recursive step. This method hinges on the assumption that if a feature is not important on its own, then it is unlikely to be part of a set of features that is important. However, this assumption may fail in several situations. 
    \end{enumerate}
    Other effective strategies can be formulated for choosing the basis lattices for different tasks and algorithms. In future work, we intend to carry out a comparative study of different strategies on different outlier detection and classification datasets.

    \subsection{Learning algorithm}\label{ssec:learning algorithm}
   Let $\mathbb{P} = (A, X, I)$ be a formal context. Once the collection $\mathcal{T}$ of  designated (crisp) agendas 
   has been fixed, we can define a meta-learning algorithm to learn an  appropriate fuzzy agenda for 
   the given  outlier detection or classification algorithm from the space of mass functions on these agendas (i.e.~fuzzy agendas).  Let $\mathcal{T} \subseteq \mathcal{P}(X)$ be the collection of designated crisp agendas, and let $Alg$ be an FCA-based  outlier detection (resp.~classification) algorithm assigning an outlier degree to every element of set  $A$ (resp.~classifying the objects in $A$  into $n$ classes) for a given agenda $Y$. For any object $a$ and a class $k$, let $Alg(a, Y)$ (resp.~$Alg_k(a, Y)$) denote the outlier degree (resp.~degree of membership of  object $a$ into  class $k$) according to the outlier detection (resp.~classification) algorithm $Alg$ acting on the context $ \mathbb{P}_Y$ for some $Y \in \mathcal{T}$. Notice that we allow for the  outlier detection algorithms (resp.~classifiers)  to be  fuzzy or probabilistic in the sense that the outlier degree (resp.~membership value to some class) of $a$ assigned by an algorithm  belongs to $[0,1]$. 
   
For a given weight function $w: \mathcal{T} \to \mathbb{R}$, the outlier degree of $a$ (resp.~the membership of $a$ in the class $k$) assigned by the algorithm $Alg$ acting on a non-crisp categorization described by $w$ is 
    \begin{equation}
    \label{eqn:outputs}
    outdeg(a) =  Alg(a,w) = \frac{\sum_{Y\in \mathcal{T} } w(Y)Alg(a, Y)}{\sum_{Y \in \mathcal{R} }w(Y)}.
    \end{equation}
    \begin{equation}
    \label{eqn:outputs 2}
    Alg_k(a,w) = \frac{\sum_{Y\in \mathcal{T} } w(Y)Alg_k(a, Y)}{\sum_{Y \in \mathcal{T} }w(Y)}.
    \end{equation}
    Intuitively, this corresponds to taking the weighted sum of the result given by $Alg$  w.r.t.~different agendas with  the weights defined by the agenda $w$. Let  $loss_{out}$ (resp.~$loss_{class}$) be the  loss functions for a given outlier (resp.~classification) task. We use a gradient descent method to learn the appropriate weight function $w_0$  that minimizes this loss. This weight function is then used  to assign an outlier degree or a classification  class to an object. That is,   for any test object $b$, its predicted outlier degree (resp.~membership in class $k$) is given by $ Alg(b,w_0) $ (resp. $Alg_k(b, w_0)$). 
    
\begin{algorithm}
\footnotesize
\caption{Meta-Learning Algorithm for Interrogative Agendas}
\label{alg:meta-learning}
\hspace*{\algorithmicindent} \textbf{Input:} a set of objects $A$, a set of features $X$, a training set $Train \subseteq A$, and a map $y:Train \to C$ representing the labels on the training set, an algorithm $Alg$ that takes in input some object and a formal context $\mathbb{P}_Y$ for any $Y \in \mathcal{T}$, and outputs an outlier degree of that object w.r.t.~that concept lattice; a loss function $loss$ for outlier detection, and a number of training epochs $M$.\\
\hspace*{\algorithmicindent} \textbf{Output} A model that predicts outlier degree of every object $a \in A$.
\begin{algorithmic}[1]
\Procedure{Train}{$A$, $X$, $Train$, $y$, $Alg$, $loss$,  $M$}
    \State \textbf{compute}  the results of the algorithm $Alg$, $Alg(a,Y)$ w.r.t.~all the agendas $Y \in \mathcal{T}$ for all objects $a \in A$
    \State \textbf{let} $outdeg$ be an empty map from $A$ to $\mathbb{R}^C$
    \State \textbf{let} $w$ be an array of weights of length $n$ initialized with random values in $\mathbb{R}$
    \For{$e = 1, \ldots, M$ } 
        \For{$a \in X$, $k \in C$}
            \State $outdeg[a] \leftarrow \frac{\sum_{Y \in \mathcal{T}} w(Y)Alg(a,Y)}{\sum_{Y \in \mathcal{T}} w(Y)}$
        \EndFor
        \State \textbf{update} $w$ with an iteration of gradient descent using $loss(predictions)$
    \EndFor
\EndProcedure
\end{algorithmic}
\end{algorithm}
{A generic  algorithm for classification can be given in a similar manner. By taking  $Alg$ to be the unsupervised algorithm discussed in Section \ref{sec:unsupervised}, we obtain a {\em supervised algorithm}, which is the main proposal of this paper for a supervised outlier detection algorithm.   
}

\section{Performance of the algorithms} \label{sec:performace}
Throughout this section, for any formal context $\mathbb{P} = (A, X, I)$, we set the space of designated crisp agendas (cf.~Section \ref{ssec:space of possible agendas}) to be  $\mathcal{T}: = \{X\}\cup\{Y\subseteq X\mid |Y|\leq 2\}$.  Thus, for any object $a$, the outlier degree of $a$ assigned by the unsupervised algorithm is given by $outdeg(a, \mathcal{T})$, where $\mathcal{T}$
is  defined as above. 
%
We  choose the parameter $\gamma$ randomly (cf.~Footnote \ref{footn:gamma}) from the interval $[0,1]$.  We evaluate the unsupervised and supervised algorithms introduced in the previous sections  on  the  datasets, commonly used   in the literature on outlier detection \cite{Campos2016,dal2015calibrating,dal2014learned,lereproducible,lebichot2020incremental,Rayana:2016},  listed in Table \ref{tab:datasets}.   
\begin{table}[]
    \centering
    \begin{tabular}{|c| c| c| c|}
    \hline
         Dataset &  No. of entries & No. of attributes & No. of outliers\\
         \hline
         lympho & 148 & 18 & 6 (4.1 \%) \\
        WBC & 278 & 30 & 21 (5.6 \%) \\ 
        glass & 214 & 9 & 9 (4.2 \%) \\
        vowels & 1456 & 12 & 50 (3.4 \%) \\
        cardio & 1831 & 21 & 176 (9.6 \%)\\
        thyroid & 3772 & 6 & 93 (2.5 \%)\\
        letter & 1600 & 32 & 100 (6.25 \%)\\
         pima & 768 & 8 & 268 (35 \%)\\
         satellite & 6435  & 36 & 2036 (32 \%)\\
         shuttle & 49097 & 9 & 3511 (7 \%) \\
         breastW & 683 & 9 & 239 (35 \%) \\
          ionosphere & 351 & 32 & 126 (36 \%)\\
        http & 567479 & 3 & 2211 (0.4 \%)\\
        forest cover & 286048 & 10 & 2747 (0.9 \%)\\
        smtp & 95156 &	3	& 30 (0.03 \%)\\
        mammogr.\ &  11183& 6 & 260 (2.32 \%)\\
        annthyroid	& 7200 & 	6	& 534 (7.42 \%)\\
        pendigits	& 6870 &	16 & 	156 (2.27 \%)\\
        wine	& 129	& 13	& 10 (7.7 \%)\\
        vertebral	& 240	& 6	 & 30 (12. 5\%)\\

Waveform & 	3,443 & 	21 &  100 (2.9 \%)\\
WBC &	454	& 	9  & 10  (2.2 \%)\\
Hepatitis	& 80 & 	13	& 19  (23.75 \%)\\
PageBlocks	& 5,473 & 10& 	560 (10.23 \%)	\\
Stamps & 	340	&  9  & 31 (9.12 \%)	\\
Wilt	& 4,839 & 5 & 	261 (5.4 \%)	 \\
Fraud & 284807 & 29 & 492 (0.17 \%) \\

        \hline
    \end{tabular}
    \caption{Outlier detection datasets}
    \label{tab:datasets}
\end{table}
 These datasets are first converted into formal contexts, by considering data entries as objects and attributes as features. In general, these features can take any values over a set of real numbers. Thus,  conceptual scaling (cf.~Section \ref{ssec:FCA}) over these features is necessary for the conversion of data into  a formal context. For all the datasets considered, we use {\em interval scaling} \cite{ganter2012formal} with some fixed number of disjoint intervals of equal lengths. That is, if attribute $x$ takes values over interval $[\alpha, \beta]$, then we apply interval scaling for $x$, which divides interval $[\alpha, \beta]$ into $n$ intervals given by $Int_{i \in n}=[\alpha +i*(\beta -\alpha)/n, \alpha +(i+1)*(\beta -\alpha)/n]$ for some fixed number of intervals (bins) $n$. We then run the two algorithms   on the  formal context obtained by this scaling procedure. 
In each test, we  split the datasets in a train and a test set with 0.8/0.2 proportion, stratifying on the outliers.

For the unsupervised algorithm, we performed 2-4 tests for each number $n$ of bins, where $n$ ranges in $\{5m\mid m\in \mathbb{N} \text{ and } 2\leq m\leq 20\}$. The ROC AUC  score (Henceforth referred as AUC score) for the best run, and the corresponding number of bins are reported in Table \ref{tab:unsupervised_results}, where we also report the AUC scores obtained for the same datasets using some other common unsupervised outlier detection algorithms in the survey (cf.~\cite[Table D4]{han2022adbench}).  The comparison with the unsupervised algorithms from 
DeepOD package (see \url{https://github.com/xuhongzuo/DeepOD}) is presented in the Table \ref{tab:unsupervised_results_2}. 

\begin{table}[h]
    \hspace{-0.5cm}
    \footnotesize
    \begin{tabular}{|c|c|c|c|c|c|c|c|c|c|c|c|}
        \hline
        {\bf Dataset} & \multicolumn{2}{c|}{\bf FCA uns.\ } &&&&&&&&&\\
        \cline{2-3}
        &  {\bf bins} & {\bf AUC} & {\bf KNN} & {\bf LOF} & { \bf OCSVM} & {\bf IForest} &{\bf SOD} & {\bf ECOD} &{\bf HBOS} &{\bf  CBLOF} & {\bf PCA}\\
        \hline

        lympho & 20 & 1  &  0.559  & 0.899 & 0.995 & 0.998 & 0.725 & 0.995 & 0.995& 0.998 & 0.998  \\
        WBC & 55 & 0.983 &0.906 & 0.542 & 0.990 & 0.990 & 0.946& 0.991 & 0.987 &\textbf{ 0.995} & 0.982 \\
        glass & 20 & \textbf{0.854} &0.823 & 0.692 & 0.354 & 0.771 & 0.734 & 0.757 & 0.772 & 0.829 & 0.663\\
        vowels & 20 & 0.855 &\textbf{ 0.973} & 0.931 & 0.612 & 0.739 & 0.926 & 0.458 & 0.722 & 0.899 & 0.653\\
        cardio & 20 & 0.887 & 0.766 & 0.663 & 0.939 & 0.932 & 0.732 & 0.944 & 0.847 & 0.899 & \textbf{0.955} \\
        thyroid & 85 & \textbf{0.988} &0.959 & 0.869 & 0.879 & 0.983 & 0.928 & 0.978 & 0.956 & 0.947 & 0.963 \\
        letter & 25 & 0.807 & \textbf{0.862} &0.845 & 0.462 & 0.611 & 0.841 & 0.508 & 0.597 & 0.756& 0.503 \\
        pima & 35 & 0.681 & \textbf{0.734} & 0.657 & 0.669 & 0.729 & 0.612 & 0.515 & 0.711 & 0.714 & 0.708\\
        satellite & 100 &\textbf{ 0.812} & 0.652 &0.559&0.590 &0.704&0.640 & 0.751 & 0.748 & 0.713 & 0.596   \\
        shuttle & 45 & 0.969 &0.696 & 0.571 &0.974 & \textbf{0.996}&0.695 & 0.994 & 0.986 & 0.835 & 0.986\\
        breastW & 80 & \textbf{0.993} &0.970 & 0.406 & 0.803 & 0.983 & 0.940 & 0.992 & 0.989 & 0.968 & 0.951\\
        ionosphere & 4 \footnote{The AUC score on this dataset using more than 10 bins does not surpass 0.44, hence we have made an exception and tried lower numbers of bins in this case.} & 0.892  & 0.883  & 0.906 & 0.759 & 0.845 & 0.864 & 0.756 & 0.625 & 0.907 & 0.792\\
        http & 25 & 0.977 & 0.034 & 0.275 & 0.996 & \textbf{1} & 0.780 & 0.981 & 0.995 & 0.996 & 0.997 \\ 
        forest cover & 60 & \textbf{0.982 }&0.860   & 0.846 & 0.926 & 0.867 & 0.745 & 0.934 & 0.802 & 0.893 & 0.937 \\
        smtp & 30 &\textbf{1} & 0.896 & 0.718 & 0.807& 0.897& 0.598 & 0.719 & 0.705 & 0.800& 0.884\\
        mammogr.\ & 55 & 0.897 & 0.845 &0.743 & 0.850 & 0.864 & 0.815 & \textbf{0.907} & 0.863 & 0.837 & 0.887\\
        annthyroid & 70 & \textbf{0.815} & 0.717  & 0.702 & 0.572 & 0.820 & 0.774 & 0.780 & 0.601 & 0.623 & 0.662\\
        pendigits & 80 & 0.942 &0.729  & 0.480 & 0.937 & \textbf{0.948} & 0.663 & 0.912 & 0.930 & 0.904 & 0.937 \\
        wine & 30 & \textbf{1} &0.450  & 0.377 & 0.731 & 0.804 &0.461 & 0.713 & 0.914 & 0.259 & 0.844 \\
        vertebral & 50 &\textbf{ 0.679} &0.338 &0.493 & 0.380 & 0.367 & 0.403 & 0.375 & 0.286 & 0.414 & 0.371 \\
        Waveform & 95 & \textbf{0.798 }& 0.738 &0.733 & 0.563 & 0.715 & 0.686 & 0.732 & 0.688 & 0.724 & 0.655\\
        WBC & 40 & 0.977 & 0.901  & 0.542 & 0.990 & 0.990 & 0.946 &0.991 & 0.987 & 0.995 & 0.982\\
    Hepatitis	& 55 & \textbf{0.923} & 0.528  & 0.380 & 0.677 & 0.697 & 0.682 & 0.800 & 0.798 & 0.664 & 0.759\\
    PageBlocks	& 25 &\textbf{ 0.97} &0.819  & 0.759 & 0.888& 0.896 & 0.777& 0.909 & 0.806 & 0.850 & 0.906 \\
    Stamps & 25 & \textbf{0.984} &0.686 & 0.513 & 0.839 & 0.912 & 0.733 & 0.914 & 0.907 & 0.682 & 0.915 \\
    Wilt & 95 &\textbf{ 0.665 }& 0.484 & 0.506 & 0.313 & 0.419 & 0.532 & 0.394 &  0.323 & 0.325 & 0.204\\
    Fraud\footnote{To reduce the memory requirements, only the formal contexts corresponding to single attributes have been used in this case. Using also all the pairs generally yields better results, but requires more memory, i.e., around $54\mathrm{GB}$ for $70$ bins.} & 70  & \textbf{0.950} & 0.936 & 0.949 & 0.906 & 0.904 & 0.950 & 0.898 & 0.903 & 0.917 & 0.903 \\
        \hline 
    Average &  &\textbf{0.899 }&0.731 &0.650  &0.756 & 0.829&0.745  & 0.800 & 0.794  & 0.790 &0.800 \\
    \hline
    \end{tabular}
    \caption{The number of bins and AUC score for the best run of FCA-based unsupervised algorithm compared with the commonly used unsupervised algorithms for outlier detection k-th nearest neighbour (KNN)  \cite{ramaswamy2000efficient}, Local outlier factor (LOF) \cite{brummer2012description}, One-class SVM (OCSVM) \cite{scholkopf1999support}, Isolation Forest (IForest)\cite{liu2008isolation},  Subspace Outlier Detection (SOD) \cite{kriegel2009outlier}, Empirical-Cumulative-distribution-based Outlier Detection (ECOD) \cite{li2022ecod}, Histogram- based outlier detection (HBOS) \cite{goldstein2012histogram}, Clustering Based Local Outlier Factor (CBLOF) \cite{he2003discovering}, and Principal Component Analysis (PCA) \cite{shyu2003novel} when all the labels are provided to these algorithms  as reported in \cite{han2022adbench}.}
    \label{tab:unsupervised_results}
\end{table}

\begin{table}[h]
\centering
    \hspace{-0.5cm}
    \footnotesize
    \begin{tabular}{|c|c|c|c|c|c|c|c|}
        \hline
        {\bf Dataset} & \multicolumn{2}{c|}{\bf FCA uns.\ } &&&&&\\
        \cline{2-3}
        &  {\bf bins} & {\bf AUC} & {\bf DeepSVDD} & {\bf REPEN} & { \bf RDP} & {\bf RCA}  & {\bf SLAD} \\
        \hline

        lympho & 20 & \textbf{1}  &  0.931  & 0.899 & \textbf{1} & \textbf{1}  & 0.995   \\
        WBC & 55 & 0.983 &0.906 & 0.542 & 0.990 & 0.990  & \textbf{0.991}  \\
        glass & 20 & {0.854} &0.817 & 0.939 & 0.866 & 0.634  & \textbf{0.951} \\
        vowels & 20 & 0.855 &0.576 & 0.963 & 0.844 & 0.891 &  \textbf{0.953} \\
        cardio & 20 & 0.887 & \textbf{0.919} & 0.757 & 0.756 & 0.867&  0.576 \\
        thyroid & 85 & \textbf{0.988} &0.775 & 0.938 & 0.929 & 0.810 &  0.951 \\
        letter & 25 & 0.807 & 0.673 &0.750 & 0.673 & 0.584 & \textbf{0.866} \\
        pima & 35 & 0.681 & 0.436 & \textbf{0.699} & 0.617 & 0.630 &  0.481 \\
        satellite & 100 &\textbf{ 0.812} & 0.462 &0.794&0.685 &0.592& 0.751   \\
        shuttle & 45 & 0.969 &0.315 & 0.978 &\textbf{0.992} & 0.990&  0.985\\
        breastW & 80 & \textbf{0.993} &0.785 & 0.961 & 0.957 & 0.993 &   0.742\\
        ionosphere & 4 \footnote{The AUC score on this dataset using more than 10 bins does not surpass 0.44, hence we have made an exception and tried lower numbers of bins in this case.} & 0.892  & 0.916  & 0.922 & 0.923 & 0.692 &   \textbf{0.983}\\
        http & 25 & 0.977 & 0.981 & 0.275 & 0.993 & \textbf{0.995}  & * \\ 
        forest cover & 60 & \textbf{0.982 }&0.511   & 0.846 & 0.201 & 0.422  & * \\
        smtp & 30 &\textbf{1} & 0.896 & 0.859 & 0.810& 0.897 & * \\
        mammogr.\ & 55 & 0.897 & 0.851 &\textbf{0.900} & 0.783 & 0.888  & 0.746\\
        annthyroid & 70 & 0.815&0.690 & 0.752  & 0.645 &0.602   &  \textbf{0.856} \\
        pendigits & 80 & 0.942 &0.739  &\textbf{ 0.952 }& 0.912 & 0.869 &   0.948  \\
        wine & 30 & \textbf{1} &0.208  & \textbf{1} & 0.687 & 0.979  &\textbf{1}\\
        vertebral & 50 &\textbf{ 0.679} &0.373 &0.520 & 0.393 & 0.468 &   0.278 \\
        Waveform & 95 & \textbf{0.798 }& 0.506 &0.766 & 0.580 & 0.754 &  0.685 \\
        WBC & 40 & 0.977 & 0.535 & \textbf{0.988} & 0.884 & 0.953  &0.779 \\
    Hepatitis	& 55 & {0.923} & 0.769  & 0.923 & 0.923 & 0.923 &   \textbf{1} \\
    PageBlocks	& 25 &\textbf{ 0.970} &0.855 & 0.964 & 0.870& 0.742 &  0.949  \\
    Stamps & 25 & \textbf{0.984} &0.952 &\textbf{ 0.984} & 0.935 & 0.822   & 0.919  \\
    Wilt & 95 &\textbf{ 0.665 }& 0.299 & 0.535 & 0.434 & 0.394   & 0.638 \\
    \hline
    Average &  &\textbf{0.902 }&0.680 &0.834  &0.783 & 0.784  & 0.827 \\
    \hline
    \end{tabular}
    \caption{The number of bins and AUC score for the best run of FCA-based unsupervised algorithm compared with some more  unsupervised algorithms. These algorithms are Deep Support Vector Data Description (DeepSVDD) \cite{pmlr-v80-ruff18a},
    Representations for a Random Nearest Neighbor Distance-based Method (REPEN) \cite{pang2018learning}, Random Detection Protection model (RDP) \cite{10.5555/3491440.3491848}, Robust Collaborative Autoencoders (RCA) \cite{liu2021rca}, Scale Learning-based deep Anomaly Detection method (SLAD) \cite{10.5555/3618408.3620019}. Many of these algorithms are implemented using the DeepOD package (see \url{https://github.com/xuhongzuo/DeepOD}).
    We use * to denote that the memory ran out when this algorithm was applied to the given dataset. We did not include the Fraud dataset as it ran out of the memory for the most algorithms. }
    \label{tab:unsupervised_results_2}
\end{table}

\begin{table}[]
\centering
 \footnotesize
    \begin{tabular}{|c|c|c|c|c|c|c|c|c|c|}
        \hline
        {\bf Dataset} &  \multicolumn{2}{c|}{\bf FCA sup.\ } &&&&&&&\\
        \cline{2-3}
        &{\bf bins} & {\bf AUC} & {\bf DevNet} & {\bf PReNet} & {\bf XGBOD} & {\bf SVM} & {\bf MLP} & {\bf RF}&  {\bf XGB} \\
        \hline
        lympho & 20 & 1  & 1 & 1 & 1 & 1 & 1& 1 & 1   \\
        WBC & 75 & \textbf{1} &0.996 & 0.989 & 1 & 1& 0.989& 1& 1 \\
        glass & 20 & 0.951  & 0.888 &0.904 &\textbf{ 1} & 0.995 & 0.927 & 1 & 1 \\
        vowels & 20 & 0.919  &0.990  &0.989 & 0.984 & \textbf{1} & 0.963 & 0.987 &\textbf{0.997}\\
        cardio & 20 & \textbf{0.998} &0.990 &0.989 &{ 0.998} & 0.997 & 0.994 & 0.997 & 0.995    \\
        thyroid & 70 & {0.999}  &0.998 &0.998 & {\bf 1} & 0.998 & 0.998 & { 1} & 1 \\
        letter & 95 & 0.931   & 0.990  & 0.989 & 0.984 & {\bf 1} & 0.963 & 0.987& {0.996} \\
        pima & 25 & 0.744  & 0.837 & 0.834& 0.924 & 0.871 & 0.856 & 0.961 &\textbf{ 0.933 }\\
        satellite & 100 & 0.812  & 0.840 &0.882 & 0.967 & 0.955 & 0.944 & 0.974 & \textbf{0.975} \\
        shuttle & 85 & \textbf{1}  & 0.978 & 0.976 & 1 & 0.990 & 0.982 & 1 & 1 \\
        breastW & 100 &\textbf{1}   & 0.997 & 0.998 &{ 0.999} & 0.997 & 0.998& 0.999 & 0.999\\
        ionosphere & 5 & 0.979  &0.948 &0.949 & \textbf{1} &0.989 & 0.991 & 1 & 1 \\
        http & 20 &  1& 1 & 1 & 1& 1 & 1 & 1 & 1\\ 
        forest cover & 85 &  1 & 1& 0.999 & 1 & 1 & 1 &1 & 1 \\
        smtp & 25 & 1 &0.855 & 0.789 & 0.988 & 0.841 & 0.864& 1 & 0.997 \\
        mammogr.\ & 85 & 0.943  & 0.931&0.932 & {\bf 0.956} & 0.874 & 0.952 & 0.947 & 0.936 \\
        annthyroid & 95 & 0.987  & 0.827 & 0.828 &\textbf{ 0.996} & 0.987& 0.986 & 0.996 & 0.996\\ 
        pendigits & 25 & 0.997 &0.998 &0.998 & \textbf{1} & 1& 0.999& 1 & 1  \\
        wine & 90 & 1  & 1& 1 & 1& 1  & 1& 1 & 1 \\
        vertebral & 85 & 0.579  &0.811  &0.822 & \textbf{0.996} & 0.916 & 0.869 &0.994 & 0.993  \\
        Waveform & 25 & 0.889  & 0.934 &0.940 & 0.937 & 0.954 &\textbf{0.968} & 0.935 & 0.936  \\
        WBC & 20 & \textbf{1}  & 0.995  &0.989 &\textbf{1}& \textbf{1}& 0.989 & \textbf{1} & \textbf{1} \\
    Hepatitis	& 20 &{\bf 1}  &0.991 &0.994 & 0.998 & \textbf{1} & \textbf{1} & \textbf{1} & \textbf{1}\\
    PageBlocks	& 85 & 0.968 &0.892 &0.908 & 0.990 & 0.969 & 0.973 & 0.990 & {\bf 0.991}\\
    Stamps & 20 &{0.984}  & 0.985&0.991& 0.999& 0.996 & 0.992 &\textbf{ 1} & 0.999\\
    Wilt & 55 & 0.857  & 0.682&0.687 & \textbf{0.991} & 0.992 & 0.867 &0.985 & 0.991 \\
    Fraud & 11 & \textbf{0.964} & 0.912 & 0.922 & 0.957 & 0.913 & 0.840 & 0.921 & 0.964 \\
        \hline
    Average &  &0.944 &0.936  &0.937 &0.987 &0.971  &0.959  &.988  &\textbf{0.989 }\\
    \hline
    \end{tabular}
    \caption{The number of bins and AUC score for the best run of FCA-based supervised algorithm compared with the commonly used supervised outlier detection algorithms Deviation Networks (DevNet) \cite{pang2019deep}, Pairwise Relation prediction-based ordinal regression Network (PReNet) \cite{pang2019deepweak}, Extreme Gradient Boosting Outlier Detection (XGBOD) \cite{zhao2018xgbod}, Support Vector Machine (SVM) \cite{cortes1995support}, Multi-layer Perceptron (MLP) \cite{rosenblatt1958perceptron}, Random Forest (RF)\cite{breiman2001random}, eXtreme Gradient Boosting (XGBoost) \cite{chen2016xgboost}  as reported in \cite{han2022adbench}.}
    \label{tab:supervised_results}
\end{table}

As to the supervised algorithm, recall that the outlier degree of an object $a$ corresponding to  a given weight function $w:\mathcal{T} \to [0,1]$ 
is  
\[
outdeg(a)= \frac{\sum_{Y \in \mathcal{T}} w(Y)Alg(a,Y)}{\sum_{Y \in \mathcal{T}} w(Y)}, 
\]
where $Alg$ is the algorithm discussed in  Section \ref{sec:unsupervised}. We learn the weights of different agendas  by applying  the meta-learning Algorithm \ref{alg:meta-learning} on this algorithm with  loss function 
\[
loss= \sum_{i \in O} outdeg(a)^2+ \sum_{i \not \in O} (1- outdeg(a))^2/bal,
\]
where $O$ is the set of outliers (from a given training set),   $outdeg (a)$ is the outlier degree of $a$ predicted by the algorithm in the given iteration, and $bal=$ (no.~of entries in the training set ) / (no.~of outliers in the training set). The balance term is used to correct for the imbalance between the sizes of sets of outliers and  inliers. 

 For every dataset, the data is split into the train and test datasets in the ratio of 0.8/0.2. We performed 2-4 tests for each number $n$ of bins, where $n$ ranges in $\{5m\mid m\in \mathbb{N} \text{ and } 2\leq m\leq 20\}$. The ROC AUC  score (Henceforth referred as AUC score) for the best run, and the corresponding number of bins are reported in Table \ref{tab:supervised_results}, where we also report the AUC scores obtained for the same datasets using some other common unsupervised outlier detection algorithms in the survey (cf.~\cite[Table D18]  
  {han2022adbench}).  It is clear that on average the supervised algorithm obtains higher AUC scores compared to the unsupervised algorithm. This shows that learning appropriate agendas can indeed help in improving the performance of the algorithm. 

\section{{Explainability of the algorithms}}\label{sec:explnations}
In this section, we describe how both the unsupervised and the supervised algorithms presented in this paper can  provide both the local and the global explanations for their outputs. Two different but interconnected types of questions can be asked about the output of the unsupervised and the supervised algorithm.
As to (a),  the outlier degree of $a$ w.r.t.~$Y$ is completely determined by the 
size of the smallest concept containing $a$ according to $Y$. Intuitively, this means that,  if $a$ is assigned a  high outlier degree w.r.t.~$Y$, this is because there are few other individuals similar to $a$ according to $Y$. %

As to (b), 
the outlier degree of any object $a$ is the average of the outlier degrees of $a$ w.r.t.~all agendas, each of which
provides information about the contribution of that agenda  to the overall outlier degree. Moreover, for any object $a$ with high overall outlier degree, the algorithm ranks the agendas according to their contribution. 

Thus, this algorithm does not just give results which can be readily interpreted by humans, it also allows to trace back which input data and which steps in the algorithm  have determined the final result. In other words, this algorithm can be considered  white box.


For any object $a$ which is assigned a high outlier degree, the algorithm can provide explanation of the following form: ``The algorithm assigns  outlier degree $outdeg(a)$ to $a$ which is contributed mainly by $a$ having high outlier degrees $outdeg(a, Y_1)$,   $outdeg(a, Y_2)$, $\ldots$, $outdeg(a, Y_n)$ w.r.t.~agendas $Y_1$, $Y_2,\ldots, Y_n$." As an example, if $outdeg(a, Y_1)$ is high, then  further explanation for this result can be given as follows: ``The outlier degree  $outdeg(a, Y_1)$ is inversely proportional (cf.~Definition \ref{def:outdeg(a,Y)}) to the size $\alpha$ of  the smallest concept containing $a$ in the concept lattice generated by  $Y_1$.    That is, there are only $\alpha$ elements similar to $a$ according to agenda $Y_1$." These local explanations regarding individual outliers can be aggregated by the user, so as to give global interpretation to results such as ``agenda $Y_1$ is the agenda for which $n\%$ of objects have high outlier degree". Examples for the explanations provided by the unsupervised algorithm on actual datasets  are collected in Section \ref{sssec:examples_explain_unsup}.

\subsection{Explanations for the supervised algorithm}
The supervised algorithm is obtained by using the meta-learning algorithm to learn the best (fuzzy) agenda for outlier detection by the unsupervised algorithm. 
The outlier degree for any object is then obtained by taking the weighted average of the outlier degrees w.r.t.~all the different agendas, where the weights are provided by the  learned fuzzy agenda.  
As to (a), 
the way the supervised algorithm assigns an outlier degree to an object $a$ w.r.t.~an agenda $Y$ is identical to that of the unsupervised algorithm.

As to (b), the main difference from the unsupervised algorithm is in the way the outlier degrees w.r.t.~different agendas are aggregated; in the case of the supervised algorithm the aggregation uses weighted average rather than simple average, and the weights are learned through supervision. These weights are interpretable as the relative importance of different agendas in assigning outlier degrees. However, as the weights are learned through stochastic gradient descent, they cannot be fully explained. Thus the supervised algorithm is still interpretable, but not white box.

For any object $a$ which is  assigned a high outlier degree, the algorithm can provide explanations of the following form: ``The algorithm assigns  outlier degree $outdeg(a)$ to $a$, which is contributed mainly by $a$ having high outlier degrees $outdeg(a, Y_1)$,   $outdeg(a, Y_2), \ldots, outdeg(a, Y_n)$ w.r.t.~agendas $Y_1$, $Y_2,\ldots, Y_n$ with weights $w(Y_1)$,  $w(Y_2),\ldots, w(Y_n)$ in the learned fuzzy agenda.''  The further explanations can be provided regarding each of these outlier degrees as in case of the unsupervised algorithm.  For any crisp agenda $Y_1$, and a given object $a$, the weight learned for  in the fuzzy agenda learned by the algorithm  informs  importance of outlier degree $a$  w.r.t.~$Y_1$ in the outlier degree (output) of $a$. That is, if an agenda $Y_1$ has high (resp.~low) weight assigned to it in the learned fuzzy agenda, then the outlier degree of an object $a$ w.r.t~the agenda $Y_1$ has high (resp.~low) effect on the  outlier degree of $a$.  Thus, the weights of different crisp agendas provide global explanations for the results. 
Examples of  explanations provided by the supervised algorithm on actual datasets  are collected in Section \ref{ssec:Supervised 
 explanation}.

\subsection{Examples of explanations}\label{ssec:examples_explain}
In this Section, we give examples of explanations provided by the unsupervised and supervised algorithm for the mammography dataset. The following table lists the short names used for different attributes of the dataset.  

\subsubsection{The mammography example}
The mammography dataset  \cite{woods1993comparative}
is obtained from ODDS outlier detection dataset library \cite{Rayana:2016}.  The machine-learning 
task is detect breast cancer from radiological scans, specifically the presence of clusters of microcalcifications that appear bright on a mammogram.  For outlier detection, the minority class of calcification is considered as outlier class and the non-calcification class as inliers.  
The dataset contains the following numerical attributes. 

\begin{table}[h]
\begin{tabular}{|c|l|}
\hline
{\bf Short name} & {\bf Full name}\\
\hline
$x_1$ & Area of object (in pixels) \\
\hline
$x_2$ & Average gray level of the object \\
\hline
$x_3$ & Gradient strength of the object’s perimeter pixels\\
\hline
$x_4$ & Root mean square noise fluctuation in the object\\
\hline
$x_5$ & Contrast, average gray level of the object minus the average of a two-pixel wide \\ & border surrounding the object\\
\hline
$x_6$ & A low order moment based on shape descriptor \\
\hline
\end{tabular}
\caption{Attributes short names to full names}
\end{table}

\noindent The (crisp) agendas  for categorization would be given by a single attribute, a pair of attributes or the set of all the attributes. In the outlier detection these interrogative agendas would correspond to the questions like ``is the area of the object abnormal (abnormally high/low)?", ``is this average grey level of object common for the object of a given area" and so on. The explanations of outlier detection algorithm consist in  answers to these questions but also in indications of the relative importance of different such questions at the local (importance of agendas in individual outliers) and  global (overall importance of agendas) levels. In the following parts, we refer to   the agenda consisting of all the attributes $x_1$ to $x_6$ as the full agenda. 
\subsubsection{Unsupervised algorithm}\label{sssec:examples_explain_unsup}
We consider  a run of the unsupervised algorithm on the mammography dataset with 55 bins  with AUC score  $0.879$, false positive rate $0.191$, true positive rate $0.846$, and  average outlier degree over all the test data entries    $0.052$.  The following table lists as examples, $3$  outliers and  $3$ inliers taken  from the test data, along with  $10$ (crisp) agendas w.r.t.~which these entries  are assigned the highest outlier degrees, and also  their  outlier degrees w.r.t.~these agendas. For each  entry, these $10$ agendas  are the top $10$ agendas contributing to the outlier degree of  that entry. The symbol -  denotes that the outlier degree w.r.t.~all other agendas is zero or negative (up to 4 significant digits) for that entry.

\begin{table}[h]
    \centering
   \begin{tabular}{|c|c|c|c||c|c|c|}
 \hline
  & \multicolumn{3}{c||}{Outliers} & \multicolumn{3}{c|}{Inliers}\\
 \hline
 
   \textbf{Entry}  & 3335& 5576& 1105& 3607& 7928& 1668  \\
   \hline 
   {\bf Pred.} & 0.6667 & 0.2466 & 0.04685 & 0.5455 & 0.1574 & 0 \\
   \hline
1 & $x_1$-$x_2$ (1) & $x_1$-$x_5$ (1) & full (1) & $x_1$ (1) & full (1) & - \\
2 & $x_1$-$x_5$ (1) & $x_2$-$x_5$ (1) & $x_5$-$x_6$ (0.0308) & $x_1$-$x_2$ (1) & $x_2$-$x_6$ (0.7899) & - \\
3 & $x_1$-$x_6$ (1) & full (1) & - & $x_1$-$x_3$ (1) & $x_2$-$x_4$ (0.7866) & - \\
4 & $x_2$-$x_4$ (1) & $x_5$-$x_6$ (0.9856) & - & $x_1$-$x_4$ (1) & $x_1$-$x_6$ (0.3952) & - \\
5 & $x_2$-$x_5$ (1) & $x_4$-$x_5$ (0.9658) & - & $x_1$-$x_5$ (1) & $x_2$-$x_5$ (0.2718) & - \\
6 & $x_2$-$x_6$ (1) & $x_1$-$x_6$ (0.4714) & - & $x_1$-$x_6$ (1) & $x_5$-$x_6$ (0.1241) & - \\
7 & $x_4$-$x_5$ (1) & $x_3$-$x_5$ (0.0016) & - & $x_2$-$x_4$ (1) & $x_2$-$x_3$ (0.0679) & - \\
8 & $x_4$-$x_6$ (1) & $x_1$-$x_2$ (0.0002) & - & $x_3$-$x_4$ (1) & $x_4$-$x_6$ (0.0281) & - \\
9 & $x_5$-$x_6$ (1) & - & - & $x_4$ (1) & - & - \\
10 & full (1) & - & - & $x_4$-$x_5$ (1) & - & - \\
    \hline
\end{tabular}  
    \label{tab:unsup_explain}
\end{table}

Entry 3335 is assigned a high outlier degree  by a large number of agendas. This list provides the explanation for this data entry to have high outlier degree.  
Similarly,  entry 1668 is evaluated as an inlier by all  the agendas; again, this provides explanation for it having outlier degree 0. Entry 5576 is  assigned   outlier degree  0.2466 by the algorithm, which is also high. This is mainly contributed by   agendas $x_1$-$x_5$, $x_2$-$x_5$, full, $x_5$-$x_6$, $x_4$-$x_5$, and  $x_1$-$x_6$.  Similar explanations can be provided for the remaining entries. Thus, the algorithm explains the scores it assigns by indicating the contributions made by different agendas to the overall outlier degree of each entry. 

\begin{figure}\label{fig:histo}
    \centering
    \includegraphics{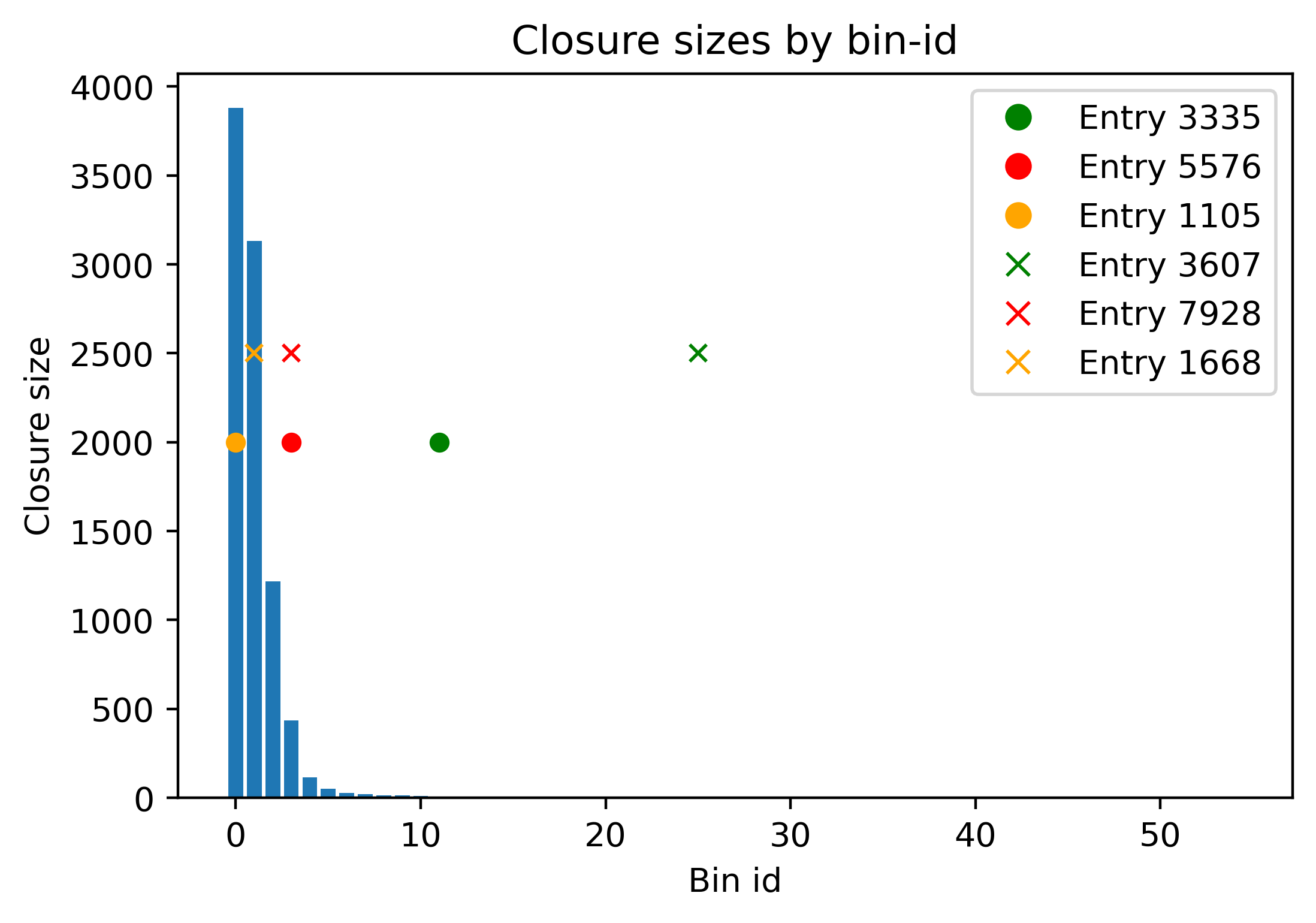}
    \caption{The histogram showing number of entries in the training data with a given closure size in the concept lattice w.r.t.~agenda $\{x_1\}$ in the training dataset. Note that  the entries whose value for attribute $x_1$  are in the same bin lie in the same (smallest) category in this lattice. Thus, closure sizes are same for all the entries with value for $x_1$ in same bin.}
    \label{fig:enter-label}
\end{figure}

A further level of explanation for these contributions can also be  provided; for  example, let us consider the case of the data entry 3607, which is assigned  outlier degree $1$  w.r.t.~agenda $x_1$. This score  can be further explained since,   at the level of granulation provided by 55 bins, there is no  entry in the train set which has a similar value  for the attribute $x_1$ (Area of object). This can be clearly seen from the histogram in Figure \ref{fig:histo}, showing the distribution of attribute $x_1$ over all the data entries in the train set. The histogram clearly  shows that the the entry 3607 has a  much higher area of shape than the mean.  On the other hand, consider  entry 1668,  which is assigned outlier degree 0 w.r.t.~agenda $x_1$ (up to four significant digits). This score can be further explained since, at the level of granulation provided by 55 bins, there are  many 
data entries  in the train set which have a similar value  for the attribute $x_1$. This can be also be clearly seen from the same histogram. 

\begin{figure}\label{fig:heatmap}
    \centering
    \includegraphics[scale = 0.7]{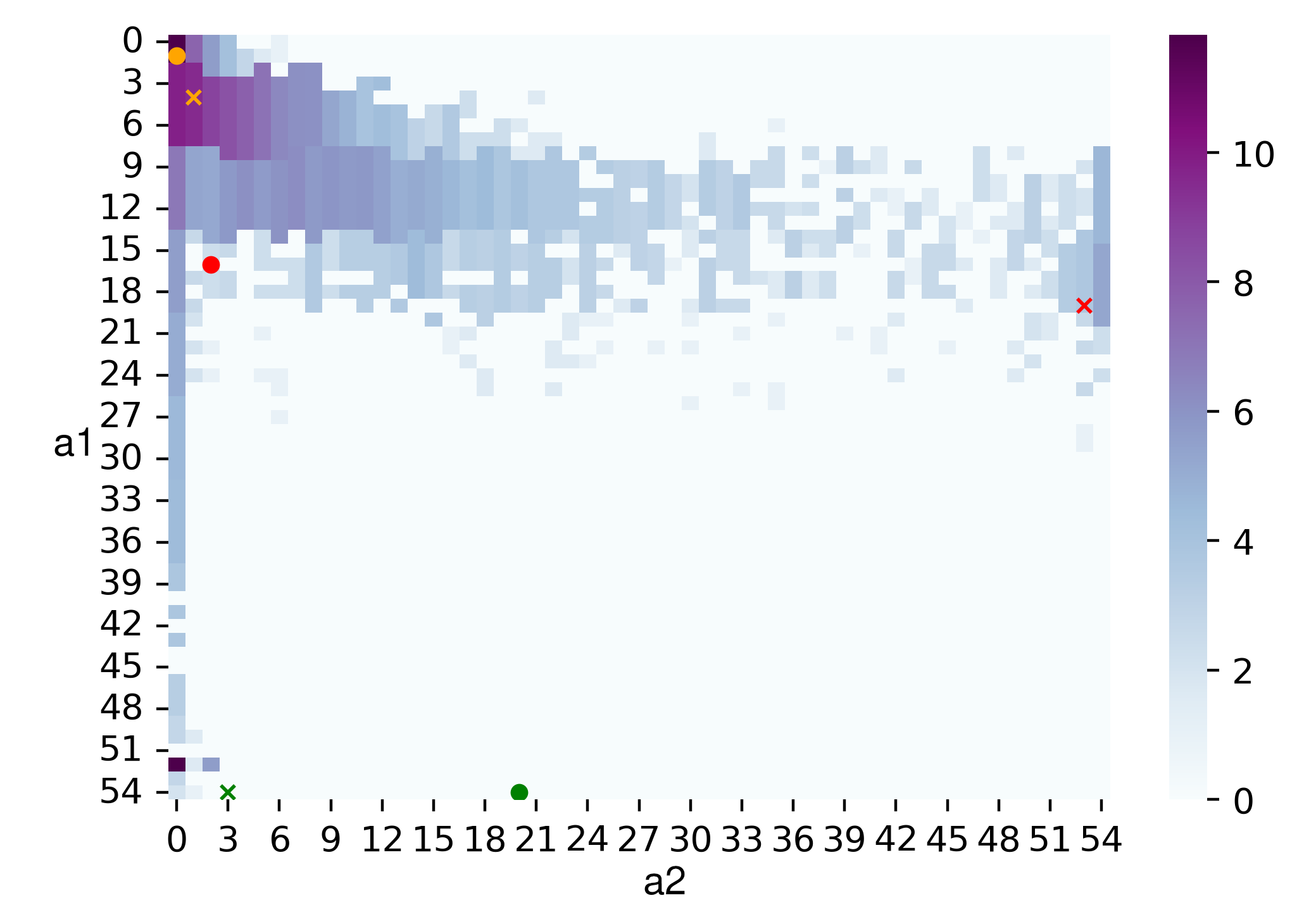}
    \caption{Heat map representing the $\log_2$ of the size of the closures w.r.t.~the agenda $\{x_1,x_2\}$ for different data entries. the legend is provided in Figure \ref{fig:histo}.
    Note that  the entries whose value for both attributes $x_1$ and $x_2$  are in the same bin lie in the same (smallest) category in this lattice. Thus, closure sizes are the same for all the entries with values for $x_1$ and $x_2$ in the same bins.}
    \label{fig:enter-label_2}
\end{figure}

We can provide the further level of explanations for the contributions made by the agendas consisting of pair of attributes in a similar manner. For example, consider the case of   entry 3335, which is assigned outlier degree 1 w.r.t.~the agenda $x_1$-$x_2$. This can be further  explained since, at the level of granulization provided by 55 bins, there is no  entry  in the  train set which has a similar values for the pair of attributes $x_1$-$x_2$ (Area of object, Average grey level of the object). This can observed from the following  heatmap in Figure \ref{fig:heatmap}. The heatmap clearly  shows that this is due to entry 3335 having much higher value for shape of object than the mean.

On the other hand, for consider entry 7928, which is assigned the outlier degree 0 w.r.t.~agenda $x_1$-$x_2$  (up to four significant digits).  This assignment   can be further  explained since,   at the level of granulation provided by 30 bins there are many  data entries  in the train set which have a similar values to the the entry 7928 for the the pair of attributes $x_1$-$x_2$. This can also be clearly seen from the same  heatmap. 

\subsubsection{Supervised algorithm}\label{ssec:Supervised 
 explanation}
We consider  a run of the supervised algorithm on the same dataset with the same number of bins and the  same values for hyper-parameter $\gamma$ in which AUC score improves to 0.926, and TPR  and FPR change to 0.788 and 0.059, respectively. The average outlier degree over all the test data entries in this run  is 0.217. The following table lists the 10 (crisp) agendas which assign the highest weighted outlier degrees with the learned weights   to same entries as in Table \ref{tab:unsup_explain}. While computing the outlier degrees w.r.t.~different agendas in the supervised algorithm, we remove the known outliers from the closures of elements. Intuitively, this corresponds to the idea that for any object $a$, it has a high (resp.~low) outlier degree, when there are small (resp.~large) number of inlier objects similar to $a$. This makes a significant difference in  the outlier degrees w.r.t.~an  agenda for some entries. For example, for the data entry 5576, the outlier degree w.r.t.~agenda $x_1$-$x_5$ increases from 0.4715 for the unsupervised algorithm to 0.7159 for the the unsupervised algorithm. 
\begin{center}
 \begin{tabular}{|c|c|c|c||c|c|c|}
  \hline
  & \multicolumn{3}{c||}{Outliers} & \multicolumn{3}{c|}{Inliers}\\ 
 \hline
   \textbf{Entry}  & 3335& 5576& 1105& 3607& 7928& 1668  \\
   \hline 
   {\bf Pred.} & 1.034 & 0.8097 & 0.4021 & 0.8836 & 0.5165 & 0 \\
   \hline
1 & full (39.3606) & full (39.3606) & full (39.3606) & full (39.3606) & full (39.3606) & - \\
2 & $x_4$-$x_5$ (16.8110) & $x_4$-$x_5$ (16.8110) & $x_5$-$x_6$ (0.4791) & $x_4$-$x_5$ (16.8110) & $x_5$-$x_6$ (1.9322) & - \\
3 & $x_5$-$x_6$ (15.5656) & $x_5$-$x_6$ (15.5656) & - & $x_4$-$x_6$ (11.2471) & $x_1$-$x_6$ (0.9728) & - \\
4 & $x_4$-$x_6$ (11.2471) & $x_1$-$x_5$ (5.2223) & - & $x_1$-$x_2$ (6.8854) & $x_4$-$x_6$ (0.3160) & - \\
5 & $x_1$-$x_2$ (6.8854) & $x_3$-$x_5$ (1.7868) & - & $x_1$-$x_5$ (5.2223) & - & - \\
6 & $x_1$-$x_5$ (5.2223) & $x_1$-$x_6$ (1.7622) & - & $x_1$-$x_4$ (2.6896) & - & - \\
7 & $x_1$-$x_4$ (2.6896) & $x_1$-$x_2$ (0.3227) & - & $x_1$-$x_6$ (2.4616) & - & - \\
8 & $x_3$-$x_5$ (2.6758) & $x_5$ (0.0175) & - & $x_4$ (2.4451) & - & - \\
9 & $x_1$-$x_6$ (2.4616) & - & - & $x_1$ (2.1649) & - & - \\
10 & $x_4$ (2.4451) & - & - & $x_1$-$x_3$ (0.4953) & - & - \\
    \hline
\end{tabular}   
\end{center}
Note that the average outlier degree increases significantly as the agenda full consisting of all the features is assigned large weight by the algorithm. This also results in a significant drop in FPR, as the data entries assigned relatively high outlier degrees in the unsupervised algorithm by contributions due to the agendas other than full  get assigned a lower relative outlier degrees in the supervised algorithm due to low relative weights assigned to these agendas. We can provide local explanations for the  outlier degrees assigned to different elements in a manner similar to that of the unsupervised algorithm. However, the contribution made to the overall outlier degree by an agenda would also depend on the weight assigned to this agenda. For example,  agenda $x_4$-$x_5$  contributes highly to the outlier degree of entry 3335. This can be further  explained since,  agenda $x_4$-$x_5$ makes a high contribution to the overall high outlier degree of  entry 3335 as at the level of granulation provided by 55 bins there are is no entry in the train data which has similar values for the pair of attributes $x_4$-$x_5$. Furthermore, this agenda is also assigned a large
weight 16.811 which makes it significant to the algorithm.

Globally, we note that the agendas full,  $x_4$-$x_5$,  $x_4$-$x_6$, and  $x_5$-$x_6$ have  significantly larger weights than other agendas. Hence, the entries which are assigned high outlier degrees w.r.t.~these agendas tend to get assigned higher outlier degrees by the algorithm. This can be interpreted as the agendas full,  $x_4$-$x_5$,  $x_4$-$x_6$, and  $x_5$-$x_6$  being given a higher importance by the supervised algorithm as compared to the other agendas. Note that  as the agenda full consists of all the features,  it assigns  high outlier degrees to a large number of entries. Thus, the difference between outlier degrees assigned to different data entries is contributed significantly by the agendas  $x_4$-$x_5$,  $x_4$-$x_6$, and  $x_5$-$x_6$.




\section{Conclusion and future directions} \label{sec:Conclusion}
In this paper, we develop an FCA-based framework for {\em explainable} outlier detection,  the explainability of which hinges on  the notion of {\em interrogative agendas}, formally represented as Dempster-Shafer mass functions. We  describe  a simple and scalable unsupervised FCA-based outlier detection algorithm, and a meta-learning algorithm aimed at identifying which  interrogative agenda gives rise to the most suitable outcome when applied in combination with  extant FCA-based categorization or outlier detection algorithms. 
Adding this learning step to any such  algorithm  allows us to improve the accuracy and sample complexity of the procedure, while also adding a level of explanability to the results. This meta-learning algorithm is then combined with the unsupervised algorithm proposed earlier, to obtain a supervised outlier detection algorithm. 
We describe the performance of both unsupervised and supervised algorithms on several commonly used datasets in outlier detection, and compare their performances  with other algorithms in the literature. 
We also describe how the unsupervised and supervised algorithms can provide both local and global explanability to their results, and, as an example, we discuss  the application of this methodology to the mammography outlier detection dataset. This work opens several interesting directions for future research.

\paragraph{Choosing agenda space.} As discussed in Section \ref{ssec:space of possible agendas}, many different strategies can be used to choose a set of crisp agendas over which the supervised algorithm learns weights. In the  future, we intend to do a comparative study of the effectiveness of these different strategies in different tasks and understanding the principles on which such choice can be based.

\paragraph{Transferability of agendas.} In the future, we intend to study  transferability analysis of   learned  agendas. That is, if an agenda learned for a certain task can be suitable for other tasks as well. In this regard, it would also be interesting to see if the agendas learned for outlier detection can be effective also for other  classification tasks. 

\paragraph{Comparison and combination of algorithms.} The agendas learned by the meta-learning algorithm can be considered as  Dempster-Shafer mass functions on the power-set of the set of attributes. Several measures of distance between mass functions have been studied  in Dempster-Shafer theory \cite{JOUSSELME2012118}. If we apply the meta-learning algorithm to different  algorithms, the distance between  agendas learned for  different algorithms can be regarded as a measure of dissimilarity between these  algorithms. Additionally, many rules for combination of evidence formulated in terms of mass functions have been developed in Dempster-Shafer theory \cite{sentz2002combination}. It would be interesting to explore the possibility of  combining agendas learned through different unsupervised algorithms for outlier detection, via different Dempster-Shafer combination rules, and using the ensuing combinations  to assemble these algorithms. As discussed in Section \ref{sec:intro},   suitable agendas for outlier detection can also be obtained by human experts. These agendas can then be  compared and combined  with the agendas learned by the machine-learning algorithms,  providing a new framework for human-machine interaction. The interaction between the  agendas learned   by the  algorithm,  and the corresponding categorizations can be studied formally using the  framework developed in \cite{FLexiblecat2022}.

\paragraph{Extending to incomplete and uncertain data.}
As discussed in Section \ref{sec:intro}, the framework of  formal concept analysis has been used to deal with uncertain or incomplete data. The algorithms developed in this paper can be generalized using these frameworks to obtain explainable outlier detection algorithms for datasets with incomplete or probabilistic data. This methodology can also be exploited for producing semi-supervised FCA-based algorithm for outlier detection.  

{\paragraph{Batching.} As the algorithms  presented in this paper use only the size of closures of an object to measure its outlier degree, these algorithms scale to large datasets much better than commonly used FCA-based outlier detection methods (cf.~Section \ref{ssec: related work}). However, the scalability to big datasets with large number of attributes 
is still limited due to the memory requirements in storing the outlier degrees w.r.t.~large number of agendas. In the future, we intend to study effective {\em batching} procedures for solving this problem.}

\paragraph{Application to  different categorization tasks.} In this paper, we have focused on the outlier detection algorithms. However, as discussed in Section \ref{sec:Learning interrogative agendas}, the meta-learning algorithm described in this paper can also be used to get explainable FCA-based algorithms for different classification tasks. In the future, we intend to carry out a similar study regarding classification. We also intend to apply the algorithms presented in this paper to outlier detection problems in domains, such as auditing, in which the legal or regulatory controls require explanations to be given  for the outcomes, and check the effectiveness of the algorithms by interacting with  domain experts.

\section*{Declaration of interest and disclaimer:} The authors report no conflicts of interest, and declare that they have no relevant or material financial interests related to the research in this paper. The authors alone are responsible for the content and writing of the paper, and the views expressed here are their personal views and do not necessarily reflect the position of their employer.
\bibliographystyle{abbrv}
\bibliography{ref}

\appendix
\end{document}